\documentclass[sigconf]{acmart}

\AtBeginDocument{%
  \providecommand\BibTeX{{%
    \normalfont B\kern-0.5em{\scshape i\kern-0.25em b}\kern-0.8em\TeX}}}

\copyrightyear{2022}
\acmYear{2022}
\setcopyright{acmcopyright}
\acmConference[WWW '22]{Proceedings of the ACM Web Conference 2022}{April 25--29, 2022}{Virtual Event, Lyon, France}
\acmBooktitle{Proceedings of the ACM Web Conference 2022 (WWW '22), April 25--29, 2022, Virtual Event, Lyon, France}
\acmPrice{15.00}
\acmDOI{10.1145/3485447.3512175}
\acmISBN{978-1-4503-9096-5/22/04}

\usepackage{xspace}
\usepackage{subcaption}
\usepackage{threeparttable}
\usepackage{enumitem}

\usepackage{amsthm}
\newtheorem{problem}{Problem}

\newtheorem{lemma}{Lemma}

\newtheorem{property}{Property}

\newtheorem{innercustomlma}{Lemma}
\newenvironment{customlma}[1]
  {\renewcommand\theinnercustomlma{#1}\innercustomlma}
  {\endinnercustomlma}

\usepackage{color}
\newcommand{\blue}[1]{{\color{blue} #1}}

\usepackage{algorithm}
\usepackage{algpseudocode}
\algrenewcommand\algorithmicrequire{\textbf{Input:}}
\algrenewcommand\algorithmicensure{\textbf{Output:}}

\newcommand{\methodone}[0]{NodeSam\xspace}
\newcommand{\methodtwo}[0]{SubMix\xspace}
\newcommand{\methodonelong}[0]{Node Split and Merge\xspace}
\newcommand{\methodtwolong}[0]{Subgraph Mix\xspace}

\begin{document}

\title{Model-Agnostic Augmentation for Accurate Graph Classification}

\author{Jaemin Yoo}
\email{jaeminyoo@snu.ac.kr}
\affiliation{%
  \institution{Seoul National University}
  \city{Seoul}
  \country{South Korea}
}

\author{Sooyeon Shim}
\email{syshim77@snu.ac.kr}
\affiliation{%
  \institution{Seoul National University}
  \city{Seoul}
  \country{South Korea}
}

\author{U Kang}
\email{ukang@snu.ac.kr}
\affiliation{%
  \institution{Seoul National University}
  \city{Seoul}
  \country{South Korea}
}

\renewcommand{\shortauthors}{Trovato and Tobin, et al.}

\begin{abstract}
Given a graph dataset, how can we augment it for accurate graph classification?
Graph augmentation is an essential strategy to improve the performance of graph-based tasks, and has been widely utilized for analyzing web and social graphs.
However, previous works for graph augmentation either a) involve the target model in the process of augmentation, losing the generalizability to other tasks, or b) rely on simple heuristics that lead to unreliable results.
In this work, we introduce five desired properties for effective augmentation. 
Then, we propose \methodone (\methodonelong) and \methodtwo (\methodtwolong), two model-agnostic algorithms for graph augmentation that satisfy all desired properties with different motivations.
\methodone makes a balanced change of the graph structure to minimize the risk of semantic change, while \methodtwo mixes random subgraphs of multiple graphs to create rich soft labels combining the evidence for different classes.
Our experiments on social networks and molecular graphs show that \methodone and \methodtwo outperform existing approaches in graph classification.

\end{abstract}





\maketitle

\section{Introduction} \label{sec:intro}

\emph{How can we augment graphs for accurate graph classification?}
Data augmentation is an essential strategy to maximize the performance of estimators by enlarging the distribution covered by training data.
The technique has been used widely in various data domains such as images \cite{DBLP:journals/jbd/ShortenK19}, time series \cite{DBLP:conf/ijcai/Wen0YSGWX21}, and language processing \cite{DBLP:conf/acl/FengGWCVMH21}.
The problem of graph augmentation has also attracted wide attention in the web domain \cite{DBLP:journals/corr/abs-2009-10564, DBLP:conf/www/Wang0LCH21}, where a community structure works as an essential evidence for classifying graph labels.
An augmentation method provides rich variants of community structures, allowing one to understand the complex relationships between users.

%The problem of graph augmentation has also attracted wide attention for graph-related tasks such as graph classification \cite{DBLP:journals/corr/abs-2009-10564} with the advancement of graph neural networks (GNN) \cite{DBLP:journals/corr/abs-1812-08434, DBLP:conf/iclr/KipfW17, DBLP:conf/iclr/XuHLJ19}.

\begin{figure}
	\vspace{2mm}
	\includegraphics[width=0.4\textwidth]{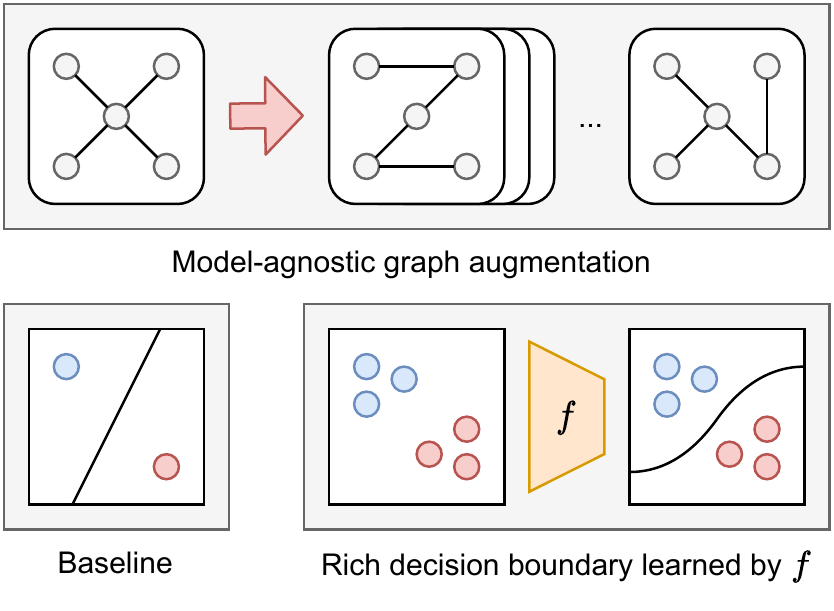}
	\vspace{-1mm}
	\caption{
		Illustration of how model-agnostic augmentation works.
		Each circle at the bottom represents a graph, whose label is represented as color.
		The augmented graphs allow a classifier $f$ to learn a rich decision boundary.
	}
	\label{fig:motivation}
\end{figure}

Previous approaches on graph augmentation are categorized to model-specific \cite{DBLP:journals/corr/abs-2006-06830, DBLP:journals/corr/abs-1909-11715} and model-agnostic ones \cite{DBLP:conf/kdd/WangWLCLH20, DBLP:journals/corr/abs-2009-10564, DBLP:conf/cikm/ZhouSX20}.
Model-specific approaches often make a better performance than model-agnostic ones, because they are designed for specific target models.
However, their performance is not generalized to other settings of models and problems, and a careful tuning of hyperparameters is required even with a small change of experimental setups.
On the other hand, model-agnostic approaches work generally well with various models and problems, even though their best performance can be worse than that of model-specific ones.
Figure \ref{fig:motivation} illustrates how model-agnostic augmentation works for a graph classifier $f$ that classifies each graph into the red or the blue class.

However, previous works on model-agnostic augmentation \cite{DBLP:conf/kdd/WangWLCLH20, DBLP:journals/corr/abs-2009-10564, DBLP:conf/cikm/ZhouSX20} rely on simple heuristics such as removing random edges or changing node attributes rather than carefully designed operations.
Such heuristics provide no theoretical guarantee for essential properties of data augmentation such as the unbiasedness or linear scalability.
As a result, previous approaches often make unreliable results, losing the main advantage over model-specific approaches.
Moreover, our experiments on benchmark datasets show that they often decrease the accuracy of target models in graph classification, where the structural characteristic of each graph plays an essential role for predicting its label (details are in Section \ref{sec:exp}).

In this work, we first propose five properties that an augmentation algorithm should satisfy to maximize its effectiveness.
These properties are designed carefully to include the degree of augmentation, the preservation of graph size, and the scalability to large graphs.
We then propose two novel algorithms, \methodone and \methodtwo, which satisfy all these properties.
\methodone (\methodonelong) performs split and merge operations on nodes, minimizing the degree of structural change while augmenting both the node- and edge-level information.
\methodtwo (\methodtwolong) combines multiple graphs by swapping random subgraphs to maximize the degree of augmentation, generating rich soft labels for classification.

%Existing methods fail to satisfy these properties, resulting in a limited performance on real-world datasets for graph classification.
%Both of these approaches achieve the highest accuracy compared with existing approaches, showing the effectiveness of augmentation for graph classification.

Our contributions are summarized as follows:
\begin{itemize} %[topsep=0pt]
	\item \textbf{Objective formulation.}
		We propose desired properties that are essential for effective graph augmentation, providing a clear objective for augmentation algorithms.
	\item \textbf{Algorithms.}
		We propose \methodone and \methodtwo, effective model-agnostic algorithms for graph augmentation.
		\methodone is a stable and balanced approach that makes a minimal change of the graph structure, while \methodtwo generates more diverse samples through abundant augmentation. 
%		\blue{\methodone aims to make a minimal change of the graphical structure, while \methodtwo aims to maximize the degree of augmentation.}
	\item \textbf{Theory.}
		We theoretically analyze the characteristics of our proposed approaches and demonstrate that they satisfy all the desired properties even in the worst cases.	
	\item \textbf{Experiments.}
		We perform experiments on nine datasets to show the effectiveness of our methods for improving graph classifiers.
		Our methods make up to 2.1$\times$ larger improvement of accuracy compared with the best competitors.
\end{itemize}

The rest of this paper is organized as follows.
In Section \ref{sec:motivation}, we define the problem of graph augmentation and present the desired properties.
In Section \ref{sec:methods}, we propose our \methodone and \methodtwo and discuss their theoretical properties.
We show experimental results in Section \ref{sec:exp} and introduce related works in Section \ref{sec:related-works}.
We conclude in Section \ref{sec:conclusion}.
All of our implementation and datasets are available at \underline{\smash{\url{https://github.com/snudatalab/GraphAug.git}}}.

\section{Problem and Desired Properties} \label{sec:motivation}

We formally define the graph augmentation problem and present desired properties for an effective augmentation algorithm.
Table \ref{table:augmentation-comparison} compares various methods for graph augmentation regarding the desired properties that we propose in this section.

\begin{table}
\centering
\caption{%
	Comparison between various approaches for graph augmentation with respect to desired properties.
	P$i$ refers to Property $i$ (see Section \ref{ssec:properties}).
	Our proposed methods satisfy all the desired properties, while the baselines do not.
}
\begin{tabular}{l|ccccc}
	\toprule
	\textbf{Method}
		& \textbf{P1} & \textbf{P2} & \textbf{P3} & \textbf{P4} & \textbf{P5} \\
	\midrule
	DropEdge \cite{DBLP:conf/iclr/RongHXH20}
		& & & & \checkmark & \checkmark \\
	GraphCrop \cite{DBLP:journals/corr/abs-2009-10564}
		& & & \checkmark & \checkmark & \checkmark \\
	NodeAug \cite{DBLP:conf/kdd/WangWLCLH20}
		& & & \checkmark & \checkmark & \checkmark \\
	MotifSwap \cite{DBLP:conf/cikm/ZhouSX20}
		& \checkmark & \checkmark & & \\
	\midrule
	\textbf{\methodone (proposed)}
		& \checkmark & \checkmark & \checkmark & \checkmark & \checkmark \\
	\textbf{\methodtwo (proposed)}
		& \checkmark & \checkmark & \checkmark & \checkmark & \checkmark \\
	\bottomrule
\end{tabular}
\label{table:augmentation-comparison}
\end{table}

\subsection{Problem Definition}

Given a set of graphs, graph augmentation is to generate a new set of graphs that have similar characteristics to the given graphs.
We give the formal definition as Problem \ref{problem:augmentation}.

\begin{problem}[Graph augmentation]
	We have a set $\mathcal{G}$ of graphs.
	Each graph $G \in \mathcal{G}$ consists of a set $\mathcal{V}$ of nodes, a set $\mathcal{E}$ of edges, and a feature matrix $\mathbf{X} \in \mathbb{R}^{|\mathcal{V}| \times d}$, where $d$ is the number of features.
	Then, the problem is to make a set $\bar{\mathcal{G}}$ of new graphs that are more suitable than $\mathcal{G}$ for the training of a model $f$, improving its performance.
\label{problem:augmentation}
\end{problem}

Although any task can benefit from graph augmentation, we use graph classification as the target task to solve by a classifier $f$.
This is because graph classification is more sensitive to the quality of augmentation than node-level tasks are, such as node classification or link prediction.
In graph classification, naive augmentation can easily decrease the accuracy of $f$ if it changes the characteristic of a graph that is essential for its classification (details in Section \ref{sec:exp}).
On the other hand, in node-level tasks such as node classification, even a simple heuristic algorithm can improve the accuracy of models by changing the local neighborhood of each target node \cite{DBLP:conf/kdd/WangWLCLH20, DBLP:conf/iclr/RongHXH20}.
Thus, the accuracy of graph classification is a suitable measure for comparing different approaches for graph augmentation.

\subsection{Desired Properties}
\label{ssec:properties}

Our goal is to generate a set of augmented graphs that maximize the performance of a graph classifier $f$ as presented in Problem \ref{problem:augmentation}.
The main difficulty of augmentation is that the \emph{semantic information} of a graph, which means the unique characteristic that determines its label, is not given clearly.
For example, in the classification task of molecular graphs, it is difficult even for domain experts to check whether an augmented graph has the same chemical property as in the original graph.
This makes it difficult for an augmentation algorithm to safely enlarge the data distribution.

We propose five desired properties for an effective augmentation algorithm to maximize the degree of augmentation while minimizing the risk of changing semantic information.
Property \ref{prop:unbiasedness} and \ref{prop:connectivity} are for preserving the basic structural information of a graph in terms of the size and connectivity, respectively.

\begin{property}[Preserving size]
	Given a graph $G = (\mathcal{V}, \mathcal{E}, \mathbf{X})$ and an augmentation function $h$, let $\bar{G} = h(G)$.
	Then, $h$ should make an unbiased change of the graph size by satisfying $\mathbb{E}[|\bar{\mathcal{V}}| - |\mathcal{V}|] = 0$ and $\mathbb{E}[|\bar{\mathcal{E}}| - |\mathcal{E}|] = 0$, where $\bar{G} = (\bar{\mathcal{V}}, \bar{\mathcal{E}}, \bar{\mathbf{X}})$.
\label{prop:unbiasedness}
\end{property}

\begin{property}[Preserving connectivity]
	Given a graph $G = (\mathcal{V}, \mathcal{E}, \mathbf{X})$ and an augmentation function $h$, let $\bar{G} = h(G)$.
	Then, $\bar{G}$ should follow the connectivity information of $G$.
	In other words, $\bar{G}$ should be connected if and only if $G$ is connected.
\label{prop:connectivity}
\end{property}

At the same time, it is necessary for an augmentation algorithm to make meaningful changes to the given graph; Property \ref{prop:unbiasedness} and \ref{prop:connectivity} are satisfied even with the identity function.
In this regard, we introduce Property \ref{prop:nodes} and \ref{prop:edges} that force an augmentation algorithm to make node- and edge-level changes at the same time. 

\begin{property}[Changing nodes]
	Given a graph $G = (\mathcal{V}, \mathcal{E}, \mathbf{X})$ and an augmentation function $h$, let $\bar{G} = h(G)$.
	Then, $h$ should make a change of nodes in $\mathcal{V}$ by satisfying either $\mathbb{E}[(|\bar{\mathcal{V}}| - |\mathcal{V}|)^2] > 0$ or $\mathbb{E}[\|\bar{\mathbf{X}} - \mathbf{X}\|_\mathrm{F}^2] > 0$, where $\bar{G} = (\bar{\mathcal{V}}, \bar{\mathcal{E}}, \bar{\mathbf{X}})$, and $\|\cdot\|_\mathrm{F}$ is the Frobenius norm of a matrix.
\label{prop:nodes}
\end{property}

\begin{property}[Changing edges]
	Given a graph $G = (\mathcal{V}, \mathcal{E}, \mathbf{X})$ and an augmentation function $h$, let $\bar{G} = h(G)$.
	Then, $h$ should make a change of $|\mathcal{E}|$, i.e., $\mathbb{E}[(|\bar{\mathcal{E}}| - |\mathcal{E}|)^2] > 0$, where $\bar{G} = (\bar{\mathcal{V}}, \bar{\mathcal{E}}, \bar{\mathbf{X}})$.
\label{prop:edges}
\end{property}

Lastly, we require the augmentation to be done in linear time with the graph size to support scalability in large graphs, which is essential for real-world applications \cite{DBLP:conf/kdd/YingHCEHL18, DBLP:conf/wsdm/JoYK18}.

\begin{property}[Linear complexity]
	Given a graph $G = (\mathcal{V}, \mathcal{E}, \mathbf{X})$ and an augmentation function $h$, let $\bar{G} = h(G)$.
	Then, the time and space complexities of $h$ for generating $\bar{G}$ should be $O(d|\mathcal{V}| + |\mathcal{E}|)$, where $d$ is the number of features, i.e., $\mathbf{X} \in \mathbb{R}^{|\mathcal{V}| \times d}$.
\label{prop:scalability}
\end{property}

We briefly review the previous approaches in Table \ref{table:augmentation-comparison} in terms of the desired properties.
DropEdge \cite{DBLP:conf/iclr/RongHXH20} and GraphCrop \cite{DBLP:journals/corr/abs-2009-10564} make a subgraph of the given graph as a result.
This makes a high risk of semantic change, as we have no clue for the essential part that determines the characteristic of the graph.
NodeAug \cite{DBLP:conf/kdd/WangWLCLH20} also changes the graph properties by adding and removing edges near a random node.
MotifSwap \cite{DBLP:conf/cikm/ZhouSX20} preserves the properties of the given graph with respect to both nodes and edges, but fails to make a sufficient amount of change.
Moreover, MotifSwap is not scalable to large graphs, as its running time is not linear with the number of edges due to the global enumeration to find all open triangles.

\section{Proposed Methods} \label{sec:methods}

\begin{figure*}
	\centering
	\includegraphics[width=0.98\textwidth]{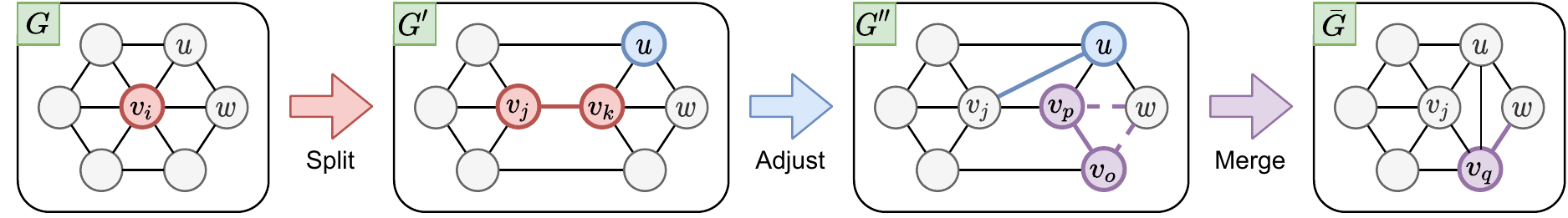}	
	\caption{
		Illustration of how the three operations of  \methodone work in an example graph: Split, Adjust, and Merge.
		The adjustment operation preserves the numbers of edges and triangles of $G$ by inserting an edge between nodes $u$ and $v_j$.
	}
\label{fig:nodesam}
\end{figure*}

In this work, we propose two effective algorithms for graph augmentation, which satisfy all the desired properties in Table \ref{table:augmentation-comparison}.
\methodone (\methodonelong) performs opposite split and merge operations over nodes to make a balanced change of graph properties, while \methodtwo (\methodtwolong) combines multiple graphs by mixing random subgraphs to enlarge the space of augmentation.

\begin{algorithm}[t]
\caption{\methodone (\methodonelong)}
\begin{algorithmic}[1]
	\Require Target graph $G = (\mathcal{V}, \mathcal{E}, \mathbf{X})$
	\Ensure Augmented graph $\bar{G} = (\bar{\mathcal{V}}, \bar{\mathcal{E}}, \bar{\mathbf{X}})$
	\State $G', v_i, v_j, v_k \gets \mathrm{Split}(G)$ \Comment{Algorithm \ref{alg:method-one-split}}
	\State $G'' \gets \mathrm{Adjust}(G, G', v_i, v_j, v_k)$ \Comment{Algorithm \ref{alg:method-one-adjust}}
	\State $\bar{G} \gets \mathrm{Merge}(G'')$ \Comment{Algorithm \ref{alg:method-one-merge}}
%	\State $y \gets 1$
%	\State $X \gets x$
%	\State $N \gets n$
%	\While{$N \neq 0$}
%	\If{$N$ is even}
%	    \State $X \gets X \times X$
%	    \State $N \gets \frac{N}{2}$  \Comment{This is a comment}
%	\ElsIf{$N$ is odd}
%	    \State $y \gets y \times X$
%	    \State $N \gets N - 1$
%	\EndIf
%	\EndWhile
\end{algorithmic}
\label{alg:method-one}
\end{algorithm}

\subsection{\methodone} \label{sec:nodesam}

The main idea of \methodone is to perform opposite operations at once to make a balanced change of the graph properties.
\methodone first splits a random node into a pair of adjacent nodes, increasing the number of nodes by one.
\methodone then merges a random pair of nodes into a single node, making the generated graph have the same number of nodes as in the original graph.
The combination of these opposite operations allows it to change the node- and edge-level information at the same time while preserving the structure of the original graph such as the connectivity.

Algorithm \ref{alg:method-one} describes the process of \methodone, where the split and merge operations are done at lines 1 and 3, respectively.
The adjustment operation in line 2 is introduced to make an unbiased change of the number of edges by inserting additional edges to the graph.
We first discuss the split and merge operations in detail and then present the adjustment operation in Section \ref{sssec:adjustment}.

%The main idea of \methodone is to perform two opposite operations at once: to split a node into a pair of nodes and to merge a pair of nodes into a single node.
%These operations preserve the structural information of the given graph, while augmenting both the nodes and edges at the same time.
%These operations make an augmentation regarding both nodes and edges while preserving the connectivity of the original graph.

Given a graph $G = (\mathcal{V}, \mathcal{E}, \mathbf{X})$, \methodone performs the split and merge operations following Algorithms \ref{alg:method-one-split} and \ref{alg:method-one-merge}, respectively.
The split operation selects a random node $v_i$ and splits it into nodes $v_j$ and $v_k$, making an edge $(v_j, v_k)$ and copying its feature as $\mathbf{x}_i = \mathbf{x}_j = \mathbf{x}_k$.
The edges attached to $v_i$ are split into $v_j$ and $v_k$ following the binomial distribution $\mathcal{B}(|\mathcal{N}_i|, 0.5)$, where $\mathcal{N}_i$ is the set of neighbors of $v_i$.
The merge operation selects a random pair of adjacent nodes $v_o$ and $v_p$ and merges them into a new single node $v_q$ with a feature vector $\mathbf{x}_q = (\mathbf{x}_o + \mathbf{x}_p) / 2$.
The edges connected to either $v_o$ or $v_p$ are connected to $v_q$, while the edge $(v_o, v_p)$ is removed.

\begin{algorithm}[t]
\caption{Split in \methodone}
\begin{algorithmic}[1]
	\Require Target graph $G = (\mathcal{V}, \mathcal{E}, \mathbf{X})$
	\Ensure Intermediate graph $G' = (\mathcal{V}', \mathcal{E}', \mathbf{X}')$, target node $v_i$, and generated nodes $v_j$ and $v_k$
	\State $v_i\gets $ Select a node from $\mathcal{V}$ uniformly at random
	\State $v_j, v_k \gets$ Make new nodes to insert to $G$
	\State $\mathbf{x}_j, \mathbf{x}_k \gets$ Make new features such that $\mathbf{x}_i = \mathbf{x}_j = \mathbf{x}_k$
	\State $h \gets$ Make a function that randomly returns $v_j$ or $v_k$
	\State $\mathcal{V}' \gets (\mathcal{V} \setminus \{v_i\}) \cup \{v_j, v_k\} $
	\State $\mathcal{E}' \gets \{ (a, b) \mid (a, b) \in \mathcal{E} \land a \neq v_i \land b \neq v_i\} \cup \{ (v_j, v_k) \}$
	\State $\mathcal{E}' \gets \mathcal{E}' \cup \{ (h(a), b) \mid (a, b) \in \mathcal{E} \land a = v_i \}$
	\State $\mathcal{E}' \gets \mathcal{E}' \cup \{ (a, h(b)) \mid (a, b) \in \mathcal{E} \land b = v_i \}$
	\State $\mathbf{X}' \gets$ Remove $\mathbf{x}_i$ from $\mathbf{X}$, and add $\mathbf{x}_j$ and $\mathbf{x}_k$ to it
\end{algorithmic}
\label{alg:method-one-split}
\end{algorithm}

\begin{algorithm}[t]
\caption{Merge in \methodone}
\begin{algorithmic}[1]
	\Require Graph $G'' = (\mathcal{V}'', \mathcal{E}'', \mathbf{X}'')$ generated from Adjust
	\Ensure Augmented graph $\bar{G} = (\bar{\mathcal{V}}, \bar{\mathcal{E}}, \bar{\mathbf{X}})$
	\State $v_o, v_p \gets $ Select adjacent nodes from $\mathcal{V}''$ uniformly at random
	\State $v_q \gets$ Make a new node to insert to $G''$
	\State $\mathbf{x}_q \gets$ Make a new feature such that $\mathbf{x}_q = (\mathbf{x}_o + \mathbf{x}_p) / 2$
	\State $\bar{\mathcal{V}} \gets (\mathcal{V}'' \setminus \{v_o, v_p\}) \cup \{v_q\} $
	\State $\bar{\mathcal{E}} \gets$ Remove all edges from $\mathcal{E}''$ connected to either $v_o$ or $v_p$
	\State $\bar{\mathcal{E}} \gets \bar{\mathcal{E}} \cup \{ (v_q, b) \mid (a, b) \in \mathcal{E}'' \land a \in \{v_o, v_p\} \}$
	\State $\bar{\mathcal{E}} \gets \bar{\mathcal{E}} \cup \{ (a, v_q) \mid (a, b) \in \mathcal{E}'' \land b \in \{v_o, v_p\} \}$
	\State $\bar{\mathbf{X}} \gets$ Remove $\mathbf{x}_o$ and $\mathbf{x}_p$ from $\mathbf{X}''$, and add $\mathbf{x}_q$ to it
\end{algorithmic}
\label{alg:method-one-merge}
\end{algorithm}

\subsubsection{Adjustment Operation}
\label{sssec:adjustment}

The basic version of \methodone with only the split and merge operations have two limitations.
First, the split operation weakens the relationships between the nodes in $\mathcal{N}_i$.
That is, the number of common neighbors between any two nodes in $\mathcal{N}_i$ is likely to decrease in the graph $G'$ generated from the split.
This happens if $v_i$ forms triangles with its neighbors, since the split eliminates these triangles by changing them into loops of length four.
Second, the number of edges tends to decrease in augmented graphs, since the merge operation can remove more than one edge.
This happens if there are triangles containing both of the target nodes $v_o$ and $v_p$, since the other two edges except $(v_o, v_p)$ in each triangle are combined into a single one.

To address the two limitations, we propose an adjustment operation of Algorithm \ref{alg:method-one-adjust} that inserts additional edges to nodes $v_j$ and $v_k$, which are generated from the split.
First, we randomly select a subset $\mathcal{S}$ of nodes from $\mathcal{T}_i$, which is the set of all nodes that form triangles with $v_i$ in the original graph $G$.
Then, we add $|\mathcal{S}|$ edges to the graph by the following process: for each node $u \in \mathcal{S}$, we insert edge $(u, s)$ to the graph, where $s$ is $v_j$ (or $v_k$) if $u$ is connected with $v_k$ (or $v_j$).
Note that all nodes in $\mathcal{S}$ have an edge with either $v_j$ or $v_k$ before the adjustment since they are neighbors of $v_i$ in $G$.

Figure \ref{fig:nodesam} illustrates how \methodone works in an example graph of seven nodes.
\methodone first splits a random node $v_i$ into a pair of nodes $v_j$ and $v_k$, decreasing the number of triangles from six to four.
Then, the adjustment operation selects a random subset $\mathcal{S} = \{u\}$ of nodes from $\mathcal{T}_i$, which is $\mathcal{V} \setminus \{v_i\}$ in this example, and connects $u$ with $v_j$.
Lastly, the merge operation combines a random pair of nodes $v_o$ and $v_p$ into node $v_q$.
Note that the numbers of edges and triangles of $G$ are preserved in the augmented graph $\bar{G}$ even with a different structure due to edge $(u, v_j)$ made by the adjustment.

%Note that $\bar{G}$ is different from $G$ since $v_q$ has a different feature vector from $v_o$ of the original graph.

%As a result, the weakened relationships between nodes around $v_j$ and $v_k$ are addressed by the new edges.

The size of $\mathcal{S}$ is determined randomly in line 7 of Algorithm \ref{alg:method-one-adjust} following a binomial distribution whose mean is given as 	$\mathbb{E}[|\mathcal{S}|] = h_i$, where $h_i$ is a number chosen to estimate the number of edges removed by the merge operation as follows:
\begin{equation}
	h_i = \frac{1}{2}((c_i^2 + 4t_i |\mathcal{V}| - 6t_i)^{1/2} - c_i),
\label{eq:adjustment}
\end{equation}
where $t_i$ is the number of triangles containing $v_i$ in $G$, $d_i$ is the node degree of $v_i$ in $G$, and $c_i = |\mathcal{E}| - 3t_i / d_i - 2$ (see Lemma \ref{lemma:adjustment}).

%\subsubsection{Comparison with Naive Adjustment}
%\label{sssec:adjustment}

%Our adjustment operation of Algorithm \ref{alg:method-one-adjust} is not the only way to make up for the missing edges removed by the merge operation.
%The simplest approach to preserve the number of edges is to perform the adjustment after the merge operation.
%In this way, we can count the exact number of edges removed by the merge operation and add the same number of edges between any pairs of nodes during the adjustment.

\begin{algorithm}[t]
\caption{Adjust in \methodone}
\begin{algorithmic}[1]
	\Require Original graph $G = (\mathcal{V}, \mathcal{E}, \mathbf{X})$, \newline
		graph $G' = (\mathcal{V}', \mathcal{E}', \mathbf{X}')$ generated from Split, \newline
		target node $v_i \in \mathcal{V}$ of Split, and \newline
		nodes $v_j \in \mathcal{V}'$ and $v_k \in \mathcal{V}'$ generated from Split
	\Ensure Intermediate graph $G'' = (\mathcal{V}'', \mathcal{E}'', \mathbf{X}'')$
	\State $t_i \gets$ Count the number of triangles in $G$ containing $v_i$
	\State $d_i \gets$ Get the degree of $v_i$ in $G$
	\State $c_i \gets |\mathcal{E}| - 3t_i / d_i - 2$ % \Comment{$d_i$ is the degree of $v_i$ in $G$}
	\State $h_i \gets ((c_i^2 + 4t_i|\mathcal{V}| - 6t_i)^{1/2} - c_i) / 2$
	\State $b \gets$ Make a function that returns a random value in $[0, 1)$
	\State $\mathcal{T}_i \gets$ Take all nodes included in the triangles containing $v_i$
%	\State $\mathcal{T}_i \gets \mathcal{T}_i \setminus \{v_i\}$
%	\State $\mathcal{S}_i \gets \{ u \mid u \in \mathcal{T}_i \land b(u) < h_i / |\mathcal{T}_i| \}$
	\State $\mathcal{S} \gets \{ u \mid (u \in \mathcal{T}_i \setminus \{v_i\}) \land (b(u) < h_i / (|\mathcal{T}_i| - 1)) \}$
	\State $\mathcal{V}'', \mathbf{X}'' \gets \mathcal{V}', \mathbf{X}'$
	\State $\mathcal{E}'' \gets \mathcal{E}' \cup \{ (u, v_j) \mid u \in \mathcal{S} \} \cup \{ (u, v_k) \mid u \in \mathcal{S} \}$
\end{algorithmic}
\label{alg:method-one-adjust}
\end{algorithm}

\textbf{Local estimation.}
All variables that compose $h_i$ in Equation \eqref{eq:adjustment} are computed from the direct neighborhood of $v_i$, except for $|\mathcal{V}|$ and $|\mathcal{E}|$ that are known in advance.
This allows \methodone to be run in linear time with the number of edges, supporting scalability to large real-world graphs.
At the same time, since the value of $h_i$ is positively correlated with the number $t_i$ of the triangles containing $v_i$, the adjustment effectively compensates for the triangles that are removed during the split; more triangles are likely to be removed with large $t_i$, but it tends to insert more edges in the adjustment.
As a result, we address the two limitations of the naive version of \methodone with a carefully chosen value of $h_i$.

\subsubsection{Satisfying Desired Properties}
\label{sssec:nodesam-properties}

\methodone satisfies all the desired properties in Table \ref{table:augmentation-comparison}.
It is straightforward that Property \ref{prop:nodes} and \ref{prop:edges} are satisfied since \methodone changes the node features and the set of edges at every augmentation.
We show in Lemma \ref{lemma:adjustment}, \ref{lemma:connectivity}, and \ref{lemma:complexity-1} that \methodone also satisfies the rest of the properties.

\begin{lemma}
	Given a graph $G = (\mathcal{V}, \mathcal{E}, \mathbf{X})$, let $\bar{G} = (\bar{\mathcal{V}}, \bar{\mathcal{E}}, \bar{\mathbf{X}})$ be the result of $\methodone$.
	Then, $\mathbb{E}[|\bar{\mathcal{V}}| - |\mathcal{V}|] = 0$ and $\mathbb{E}[|\bar{\mathcal{E}}| - |\mathcal{E}|] = 0$.
\label{lemma:adjustment}
\end{lemma}

\begin{proof}
\vspace{-1mm}
	The proof is straightforward for $\mathcal{V}$, since the number of nodes does not change by \methodone.
	The proof for $\mathcal{E}$ requires a series of estimations for the properties of intermediate graphs, and thus the full proof is given in Appendix \ref{appendix:proof}.
	The idea is that the expected number of edges removed by the merge operation is the same as $h_i$, which is the expected number of edges added by the adjustment operation as shown in line 4 of Algorithm \ref{alg:method-one-adjust}.
\vspace{-1mm}
\end{proof}

\begin{lemma}
	Given a graph $G = (\mathcal{V}, \mathcal{E}, \mathbf{X})$, let $\bar{G} = (\bar{\mathcal{V}}, \bar{\mathcal{E}}, \mathbf{X})$ be a graph generated by \methodone.
	Then, $\bar{G}$ is connected if and only if $G$ is connected.
\label{lemma:connectivity}
\end{lemma}

\begin{proof}
\vspace{-1mm}
	The proof is in Appendix \ref{appendix:proof-lemma-2}.
\vspace{-1mm}
\end{proof}

%\begin{proof}
%\vspace{-1mm}
%	The split and merge operations preserve the connectivity, since they are transformations between a single node and a pair of adjacent nodes.
%	The adjustment also preserves the connectivity since it makes new edges only between two-hop neighbors.
%\vspace{-1mm}
%\end{proof}

\begin{lemma}
	Given a graph $G = (\mathcal{V}, \mathcal{E}, \mathbf{X})$, the time and space complexities of \methodone are both $O(d|\mathcal{V}| + |\mathcal{E}|)$, where $d$ is the number of features.
\label{lemma:complexity-1}
\end{lemma}

\begin{proof}
\vspace{-1mm}
	The proof is in Appendix \ref{appendix:proof-lemma-3}.
\vspace{-1mm}
\end{proof}

\subsection{\methodtwo} \label{sec:submix}

\begin{figure}
	\includegraphics[width=0.38\textwidth]{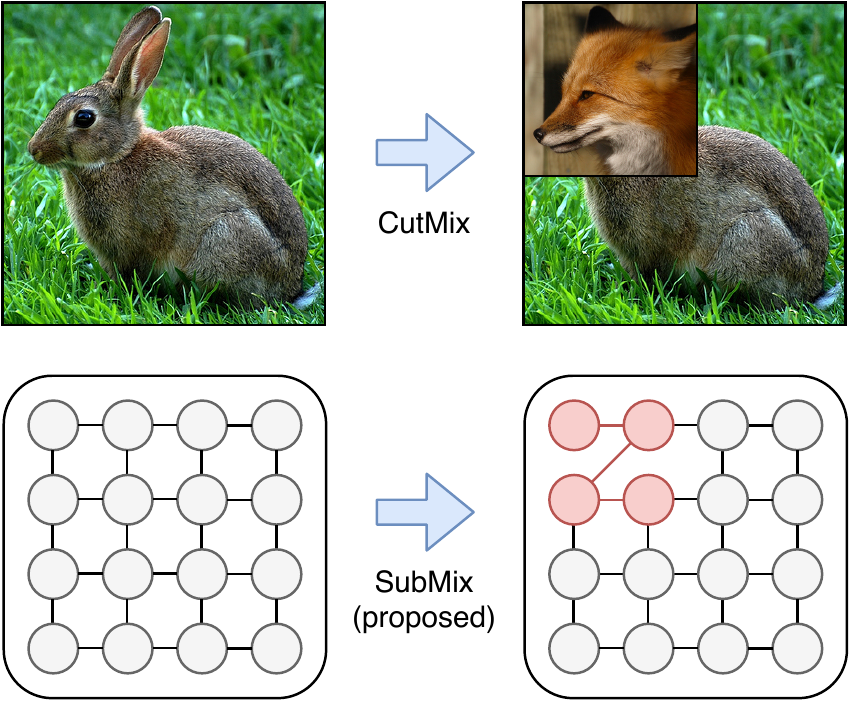}
	\vspace{1mm}
	\caption{
		Comparison between image and graph mixing augmentation.
		SubMix, our proposed approach, generalizes CutMix \cite{DBLP:conf/cikm/DuT19} to the graph domain.
		The image patch in the top row corresponds to the red inserted subgraph in the bottom.
	}
	\label{fig:submix}
\end{figure}

\methodtwo aims to make a large degree of augmentation by combining multiple graphs.
This is done by swapping subgraphs of different graphs, motivated by mixing approaches in the image domain \cite{DBLP:conf/iclr/ZhangCDL18, DBLP:conf/iccv/YunHCOYC19, DBLP:conf/icml/KimCS20}.
The main idea is to treat the adjacency matrix of each graph like an image and replace a random patch of the matrix with that of another graph, as depicted in Figure \ref{fig:submix} that compares our \methodtwo to CutMix \cite{DBLP:conf/iccv/YunHCOYC19} in the image domain.
The red subgraph corresponds to the head of the fox in the augmented image.

The overall process of \methodtwo is shown as Algorithm \ref{alg:method-two}.
Given a graph $G = (\mathcal{V}, \mathcal{E}, \mathbf{X})$ and a set $\mathcal{G}$ of all available graphs, \methodtwo makes an augmented graph by replacing a subgraph of $G$ with that of another graph $G'$ chosen from $\mathcal{G} \setminus \{G\}$.
First, \methodtwo samples ordered sets $S$ and $S'$ of nodes from $G$ and $G'$, respectively, where $|S| = |S'|$.
Then, \methodtwo makes a one-to-one mapping $\phi$ from the nodes in $S'$ to those in $S$ based on their order in each ordered set, and uses it to transfer the induced subgraph of $S'$ into $G$, replacing the induced subgraph of $S$.
The edges that connect $S$ to the rest of the graph $G$ are then connected to the new nodes.

As a result, \methodtwo makes the set $\bar{\mathcal{E}} = \mathcal{E}_1 \cup \mathcal{E}_2$ of edges for the augmented graph $\bar{G}$, where $\mathcal{E}_1$ and $\mathcal{E}_2$ are extracted from $G$ and $G'$, respectively.
The difference between $\mathcal{E}_1$ and $\mathcal{E}_2$ is that $\mathcal{E}_1$ contains the edges incident to at least one node in $S$, while $\mathcal{E}_2$ contains only the edges whose two connected nodes are both in $S'$.
The reason is because the main target of augmentation is $G$.
The subgraph of $G'$ is inserted into $G$ without changing the edges that connect $S$ with $\mathcal{V} \setminus S$.
See Figure \ref{fig:submix} for a visual illustration.

One notable difference from \methodone is that \methodtwo takes the label of $G$ as an additional input and changes it.
The label $\bar{\mathbf{y}}$ of $\bar{G}$ is softly determined as $\bar{\mathbf{y}} = q \mathbf{y} + (1 - q) \mathbf{y}'$, where $\mathbf{y}$ and $\mathbf{y}'$ are the one-hot labels of $G$ and $G'$, respectively, and $q = |\mathcal{E}_1| / |\bar{\mathcal{E}}|$.
This is based on an assumption that edges are crucial factors to determine the label of a graph, and thus the ratio of included edges quantifies how much it contributes to making $\bar{\mathbf{y}}$.

\subsubsection{Selecting Subgraphs with Diffusion}

The core part of \methodtwo is the choice of ordered sets $S$ and $S'$.
The list of nodes included in each set determines the shape of the subgraph, and their order determines how the nodes in $S'$ are connected to the nodes in $\mathcal{V} \setminus S$.
A naive approach is to select a random subset from each graph without considering the structural information, but this is likely to make disconnected subgraphs containing few edges.

Instead, we utilize graph diffusion to select connected and clustered subgraphs from $G$ and $G'$.
The process of subgraph sampling is shown as Algorithm \ref{alg:method-two-sample}.
Given a root node $r$, a diffusion operator propagates a signal from $r$ to all other nodes following the graph structure.
This assigns large affinity scores to the nodes close to $r$ or having many common neighbors with $r$.
Thus, selecting the top $k$ nodes having the largest affinity scores provides a meaningful substructure around $r$ containing a sufficient number of edges, and the chosen nodes can be used directly as the subset $S$.

We use personalized PageRank (PPR) as the diffusion function, which is a popular approach to measure the personalized scores of nodes.
PPR \cite{DBLP:conf/nips/KlicperaWG19} converts the adjacency matrix $\mathbf{A}$ of each graph to a matrix $\mathbf{S}$ of scores by diffusing the graph signals from all nodes: $\mathbf{S} = \sum_{k=0}^\infty \alpha (1 - \alpha)^k (\mathbf{D}^{-1/2} \mathbf{A} \mathbf{D}^{-1/2})^k$, where $\mathbf{D}$ is the degree matrix such that $D_{ii} = \sum_k A_{ik}$, and $\alpha = 0.15$ is the teleport probability.
The $i$-th column of $\mathbf{S}$ contains the affinity scores of nodes with respect to node $i$.
Thus, given a root node $r$, we return the $r$-th column of $\mathbf{S}$ as the result $\mathbf{c}$ of diffusion in line 3 of Algorithm \ref{alg:method-two-sample}.

\begin{algorithm}[t]
\caption{\methodtwo (\methodtwolong)}
\begin{algorithmic}[1]
	\Require
		Target graph $G = (\mathcal{V}, \mathcal{E}, \mathbf{X})$ with its label $y$, \newline
		set $\mathcal{G}$ of all available graphs with labels $\mathcal{Y}$, and \newline	
		target ratio $p \in (0, 1)$ of augmentation
	\Ensure Augmented graph $\bar{G} = (\bar{\mathcal{V}}, \bar{\mathcal{E}}, \bar{\mathbf{X}})$ and its label $\bar{\mathbf{y}}$
	\State $G', y' \gets$ Pick a random graph from $\mathcal{G} \setminus \{G\}$ and its label
	\State $S, S' \gets \mathrm{Sample}(G, G', p)$ \Comment{Algorithm \ref{alg:method-two-sample}}
	\State $\phi \gets$ Make the one-to-one mapping from $S'$ to $S$
	\State $\mathcal{E}_1 \gets \{ (u, v) \mid (u, v) \in \mathcal{E} \land \lnot (u \in S \land v \in S) \}$
	\State $\mathcal{E}_2 \gets \{ (\phi(u), \phi(v)) \mid (u, v) \in \mathcal{E}' \land (u \in S' \land v \in S') \}$
	\State $\bar{\mathcal{V}}, \bar{\mathcal{E}}, \bar{\mathbf{X}} \gets \mathcal{V}, \mathcal{E}_1 \cup \mathcal{E}_2, \mathbf{X}$
	\State $\bar{\mathbf{X}}[\phi(S')] \gets \mathbf{X}'[S']$ \Comment{Replace a subset of features}
	\State $\mathbf{y}, \mathbf{y'} \gets$ Represent $y$ and $y'$ as one-hot vectors, respectively
	\State $\bar{\mathbf{y}} \gets (|\mathcal{E}_1|/|\bar{\mathcal{E}}|) \mathbf{y} + (1 - |\mathcal{E}_1|/|\bar{\mathcal{E}}|) \mathbf{y}'$
\end{algorithmic}
\label{alg:method-two}
\end{algorithm}

\begin{algorithm}[t]
\caption{Sample in \methodtwo}
\begin{algorithmic}[1]
	\Require Target graph $G = (\mathcal{V}, \mathcal{E}, \mathbf{X})$, \newline
		another graph $G' = (\mathcal{V}', \mathcal{E}', \mathbf{X}')$ selected from $\mathcal{G} \setminus \{G\}$, and \newline
		target ratio $p \in (0, 1)$ of augmentation
	\Ensure Ordered sets $S \subseteq \mathcal{V}$ and $S' \subseteq \mathcal{V}'$ of connected nodes
	\State $r, r' \gets$ Pick random nodes from $G$ and $G'$, respectively
	\State $\psi \gets$ Make a function that finds the connected component
	\State $k \gets \mathrm{uniform}(0, p) \cdot \min(|\psi(G, r)|, |\psi(G', r')|)$
	\For{$(G_t, r_t) \in \{(G, r), (G', r')\}$}
		\State $\mathbf{c}_t \gets$ Compute the scores of nodes by $\mathrm{Diffuse}(G_t, r_t)$
		\State $S_t \gets$ Select $k$ nodes having the largest scores in $\mathbf{c}_t$
		\If{$S_t$ contains a disconnected node}
			\State $S_t \gets$ Take the first $k$ nodes from $\mathrm{BFS}(G_t, r_t)$
		\EndIf
		\State $S \gets S_t$ if $G_t = G$ otherwise $S' \gets S_t$
	\EndFor
\end{algorithmic}
\label{alg:method-two-sample}
\end{algorithm}

\textbf{Connected subgraphs.}
In Algorithm \ref{alg:method-two-sample}, it is essential to guarantee the connectivity of nodes $S$ and $S'$ to allow \methodtwo to replace meaningful substructures.
We guarantee the connectivity with two ideas in the algorithm.
First, we bound the number of selected nodes by the size of the connected component containing each root node, since the input graphs can be disconnected.
Second, if the nodes selected by PPR make disconnected subgraphs, we resample the nodes by the breadth-first search (BFS) that is guaranteed to make connected subgraphs.
This is to prevent rare cases where PPR returns a disconnected graph, which has not occurred in all of our experiments but can happen theoretically.

\begin{lemma}
	Each set of nodes returned by Algorithm \ref{alg:method-two-sample} contains only connected nodes for both $G$ and $G'$ and for any value of $k$.
\label{lemma:method-two-connectivity}
\end{lemma}

\begin{proof}
\vspace{-1mm}
	The proof is straightforward as we run BFS on the connected component of each root if $S$ (or $S'$) is not connected.
\vspace{-1mm}
\end{proof}

\textbf{Order of nodes.}
The order of selected nodes determines how the nodes in $S$ are matched with those in $S'$, playing an essential role for the result of augmentation.
One advantage of diffusion is that the selected nodes are ordered by their affinity scores, making nodes at the same relative position around the root nodes $r$ and $r'$ to be matched between $S$ and $S'$ in the replacement.
The root $r$ of $S$ is always replaced with $r'$ of $S'$, and the replacement of all remaining nodes is determined by their affinity scores.
This makes the generated graph $\bar{G}$ more plausible than in the naive approach that matches the nodes in $S$ randomly with those in $S'$.

\subsubsection{Satisfying Desired Properties}
\label{sssec:submix-properties}

\methodtwo satisfies all the desired properties in Table \ref{table:augmentation-comparison}.
It is straightforward that Property \ref{prop:nodes} and \ref{prop:edges} are satisfied since \methodtwo changes the node features and the set of edges at every augmentation.
We show in Lemma \ref{lemma:adjustment}, \ref{lemma:connectivity}, and \ref{lemma:complexity-1} that \methodtwo also satisfies the rest of the properties.

\textbf{Preserving size.}
Since \methodtwo combines multiple graphs, it is not possible to directly show the satisfaction of Property \ref{prop:unbiasedness} for any given graph $G$.
For instance, the number of edges always increases if $G$ is a chain graph and all other graphs in $\mathcal{G}$ are cliques.
Thus, we assume that the target graph $G$ is selected uniformly at random from $\mathcal{G}$ and prove the unbiasedness with $\mathbb{E}[|\mathcal{E}| - |\bar{\mathcal{E}}|] = 0$.

%We show in Lemma \ref{lemma:submix} that \methodtwo makes unbiased changes of the number of edges regardless of the specific algorithm for selecting $S$ and $S'$, as long as their selection schemes are the same.

\begin{lemma}
	Given a set $\mathcal{G}$ of graphs, we sample different graphs $G = (\mathcal{V}, \mathcal{E}, \mathbf{X})$ and $G' = (\mathcal{V}', \mathcal{E}', \mathbf{X}')$ from $\mathcal{G}$ uniformly at random.
	Let $\bar{G} = (\bar{\mathcal{V}}, \bar{\mathcal{E}})$ be an augmented graph generated by $\methodtwo$.
	Then, $\mathbb{E}[|\mathcal{V}| - |\bar{\mathcal{V}}|] = 0$ and $\mathbb{E}[|\mathcal{E}| - |\bar{\mathcal{E}}|] = 0$.
\label{lemma:submix-unbiasedness}
\end{lemma}

\begin{proof}
\vspace{-1mm}
	The proof is in Appendix \ref{appendix:proof-lemma-5}.
\vspace{-1mm}
\end{proof}

\textbf{Other properties.}
%We show in Lemma \ref{lemma:method-two-connectivity} and \ref{lemma:method-two-scalability} that \methodtwo satisfies also the other two properties.
Lemma \ref{lemma:method-two-connectivity} shows that \methodtwo preserves the connectivity of the given graph, while Lemma \ref{lemma:method-two-scalability} shows that \methodtwo runs in linear time with respect to the number of edges.

\begin{lemma}
	Given a graph $G = (\mathcal{V}, \mathcal{E}, \mathbf{X})$, let $\bar{G} = (\bar{\mathcal{V}}, \bar{\mathcal{E}}, \bar{\mathbf{X}})$ be an augmented graph generated by $\methodtwo$.
	Then, $\bar{G}$ is connected if and only if $G$ is connected.
\label{lemma:method-two-connectivity}
\end{lemma}

\begin{proof}
\vspace{-1mm}
	The proof is in Appendix \ref{appendix:proof-lemma-6}.
\vspace{-1mm}
\end{proof}

%\begin{proof}
%\vspace{-1mm}
%	Let $G'$ be a graph chosen for the augmentation of $G$ by \methodtwo.
%	The sets $S$ and $S'$ of nodes selected for the replacement are connected due to Lemma \ref{lemma:method-two-connectivity}.
%	If we treat the induced subgraphs of $S$ and $S'$ as supernodes, the replacement of $S$ with $S'$ does not change the connectivity of $G$, proving the lemma.
%\vspace{-1mm}
%\end{proof}

\begin{lemma}
	Given graphs $G = (\mathcal{V}, \mathcal{E}, \mathbf{X})$ and $G' = (\mathcal{V}', \mathcal{E}', \mathbf{X}')$, the time and space complexities of \methodtwo are $O(pd|\mathcal{V}| + |\mathcal{E}| + |\mathcal{E}'|)$, where $p$ is the ratio of sampling, and $d$ is the number of features.
\label{lemma:method-two-scalability}
\end{lemma}

\begin{proof}
\vspace{-1mm}
	The proof is in Appendix \ref{appendix:proof-lemma-7}.
\vspace{-1mm}
\end{proof}

%\begin{proof}
%\vspace{-1mm}
%	The time complexity of Algorithm \ref{alg:method-two} without including the sampling function is $O(pd|\mathcal{V}| + |\mathcal{E}| + |\mathcal{E}'|)$.
%	We assume that the number of labels is negligible.
%	The time complexity of Algorithm \ref{alg:method-two-sample} is $O(|\mathcal{E}| + |\mathcal{E}'|)$, since the PPR diffusion is $O(|\mathcal{E}|)$ and $O(|\mathcal{E}'|)$ for $G$ and $G'$, respectively.
%	The space complexities are always smaller than or equal to the time complexities \cite{DBLP:journals/jcss/Book74}.
%\vspace{-1mm}
%\end{proof}

\section{Experiments}
\label{sec:exp}

We perform experiments to answer the following questions:
\begin{itemize}
	\item[Q1.] \textbf{Accuracy (Section \ref{sec:exp-accuracy}).}
		Do \methodone and \methodtwo improve the accuracy of graph classifiers?
		Are they better than previous approaches for graph augmentation?
	\item[Q2.] \textbf{Desired properties (Section \ref{sec:exp-properties}).}
		Do \methodone and \methodtwo satisfy the desired properties of Table \ref{table:augmentation-comparison} in real-world graphs as we claim theoretically in Section \ref{sec:methods}?
	\item[Q3.] \textbf{Ablation study (Section \ref{sec:exp-ablation}).}
		Do our ideas for improving \methodone and \methodtwo, such as the adjustment or diffusion operation, increase the accuracy of graph classifiers?
\end{itemize}

\subsection{Experimental Setup}
\label{sec:exp-setup}

We introduce our experimental setup including datasets, baseline approaches, hyperparameters, and graph classifiers.

\textbf{Datasets.}
We use 9 benchmark datasets \cite{DBLP:journals/corr/abs-2007-08663, DBLP:conf/iclr/XuHLJ19} summarized in Table \ref{table:datasets}, which were used in previous works for graph classification.
D\&D, ENZYMES, MUTAG, NCI1, NCI109, PROTEINS, and PTC-MR \cite{DBLP:conf/kdd/YanardagV15} are datasets of molecular graphs that represent chemical compounds.
COLLAB \cite{DBLP:conf/kdd/YanardagV15} and Twitter \cite{DBLP:journals/tkde/PanWZ15} are datasets of social networks.
The numbers of nodes and edges in Table \ref{table:datasets} are from all the graphs in each dataset.
Detailed information of node features in the datasets are described in Appendix \ref{appendix:data}.

\begin{table}
\small
\centering
\caption{Summary of datasets.}
\begin{threeparttable}
\begin{tabular}{l|rrrrr}
	\toprule
	\textbf{Dataset} & \textbf{Graphs} & \textbf{Nodes} & \textbf{Edges} & \textbf{Features} & \textbf{Labels} \\
	\midrule
	D\&D\tnote{1} & 1,178 & 334,925 & 843,046 & 89 & 2 \\
	ENZYMES\tnote{1} & 600 & 19,580 & 37,282 & 3 & 6 \\
	MUTAG\tnote{1} & 188 & 3,371 & 3,721 & 7 & 2 \\
	NCI1\tnote{1} & 4,110 & 122,747 & 132,753 & 37 & 2 \\
	NCI109\tnote{1} & 4,127 & 122,494 & 132,604 & 38 & 2 \\
	PROTEINS\tnote{1} & 1,113 & 43,471 & 81,044 & 3 & 2 \\
	PTC-MR\tnote{1} & 344 & 4,915 & 5,054 & 18 & 2 \\
	\midrule
	COLLAB\tnote{1} & 5,000 & 372,474 & 12,286,079 & 369 & 3 \\
	Twitter\tnote{1} & 144,033 & 580,768 & 717,558 & 1,323 & 2 \\
	\bottomrule
\end{tabular}
\begin{tablenotes} \footnotesize
\item[1]\url{https://chrsmrrs.github.io/datasets}
\end{tablenotes}
\end{threeparttable}
\label{table:datasets}
\end{table}

\begin{table*}
	\small
	\centering
	\caption{
		Accuracy of graph classification with various graph augmentation methods.
		The values in parentheses are the ranks of methods in each dataset, and the Rank column shows the average and standard deviation of ranks over all datasets.
		The proposed methods \methodone and \methodtwo achieve the best average accuracy and the highest ranks at the same time.
	}
	\begin{tabular}{l|lllllllll|l|c}
		\toprule
		\textbf{Method} &
			\textbf{D\&D} &
			\textbf{ENZY.} &
			\textbf{MUTAG} &
			\textbf{NCI1} &
			\textbf{N109} &
			\textbf{PROT.} &
			\textbf{PTC-MR} &
			\textbf{COLLAB} &
			\textbf{Twitter} &
			\textbf{Average} &
			\textbf{Rank} \\
		\midrule
		Baseline &
			76.40 (4) &
			50.33 (10) &
			89.94 (4) &
			82.68 (9) &
			81.80 (9) &
			75.38 (9) &
			63.94 (7) &
			82.66 (7) &
%			91.49 (6) &
			66.05 (7) &
			74.35 (8) &
			7.33 $\pm$ 2.18 \\
		\midrule
		GraphCrop &
			77.08 (2) &
			51.00 (9) &
			77.11 (10) &
			80.46 (10) &
			79.77 (10) &
			75.20 (10) &
			61.87 (10) &
			83.50 (2) &
			66.15 (3) &
%			91.67 (3) &
			72.46 (10) &
			7.33 $\pm$ 3.77 \\
		DropEdge &
			76.14 (6) &
			53.67 (6) &
			81.93 (9) &
			82.82 (7) &
			82.60 (7) &
			75.74 (4) &
			63.68 (8) &
			82.50 (9) &
			66.05 (8) &
%			91.40 (8) &
			73.90 (9) &
			7.11 $\pm$ 1.62 \\
		NodeAug &
			76.14 (8) &
			54.67 (5) &
			86.14 (7) &
			83.16 (4) &
			82.36 (8) &
			75.56 (6) &
			66.24 (2) &
			81.32 (10) &
			65.98 (9) &
%			91.32 (10) &
			74.62 (7) &
			6.56 $\pm$ 2.55 \\
		AddEdge &
			76.14 (7) &
			55.17 (3) &
			85.67 (8) &
			83.99 (2) &
			83.06 (5) &
			75.38 (8) &
			64.27 (5) &
			82.80 (5) &
			66.10 (6) &
%			91.47 (7) &
			74.73 (6) &
			5.44 $\pm$ 2.07 \\
		ChangeAttr &
			75.72 (9) &
			53.33 (8) &
			90.44 (3) &
			83.02 (5) &
			83.57 (2) &
			75.47 (7) &
			62.47 (9) &
			82.76 (6) &
			66.36 (2) &
%			91.64 (4) &
			74.79 (5) &
			5.67 $\pm$ 2.83 \\
		DropNode &
			75.55 (10) &
			55.17 (3) &
			87.28 (6) &
			82.85 (6) &
			83.04 (6) &
			75.65 (5) &
			\textbf{66.59 (1)} &
			82.54 (8) &
			66.11 (5) &
%			\textbf{91.77 (1)} &
			74.97 (4) &
			5.56 $\pm$ 2.60 \\
		MotifSwap &
			76.23 (5) &
			53.50 (7) &
			90.47 (2) &
			82.82 (7) &
			83.28 (4) &
			75.92 (3) &
			65.79 (3) &
			82.84 (3) &
			65.93 (10) &
%			91.39 (9) &
			75.20 (3) &
			4.89 $\pm$ 2.62 \\
		\midrule
		\textbf{\methodtwo} &
			\textbf{78.10 (1)} &
			57.50 (2) &
			89.94 (4) &
			\textbf{84.33 (1)} &
			\textbf{84.37 (1)} &
			\textbf{76.19 (1)} &
			63.97 (6) &
			\textbf{83.74 (1)} &
			\textbf{66.44 (1)} &
%			91.59 (5) &
			76.06 (2) &
			\textbf{2.00 $\pm$ 1.80} \\
		\textbf{\methodone} &
			76.57 (3) &
			\textbf{60.00 (1)} &
			\textbf{90.96 (1)} &
			83.33 (3) &
			83.52 (3) &
			76.10 (2) &
			65.48 (4) &
			82.82 (4) &
			66.13 (4) &
%			91.74 (2) &
			\textbf{76.10 (1)} &
			2.78 $\pm$ 1.20 \\
		\bottomrule
	\end{tabular}	
	\label{tab:main-accuracy}
\end{table*}

\textbf{Baselines.}
We consider the following baselines for graph augmentation.
DropEdge \cite{DBLP:conf/iclr/RongHXH20} removes an edge uniformly at random, while DropNode removes a node and all connected edges.
AddEdge inserts an edge between a random pair of nodes.
ChangeAttr augments the one-hot feature vector of a random node by changing the nonzero index.
GraphCrop \cite{DBLP:journals/corr/abs-2009-10564} selects a subgraph by diffusion and uses it as an augmented graph.
NodeAug \cite{DBLP:conf/kdd/WangWLCLH20} combines ChangeAttr, DropEdge, and AddEdge to change the local neighborhood of a random target node.
MotifSwap \cite{DBLP:conf/cikm/ZhouSX20} swaps the two edges in an open triangle to preserve the connectivity during augmentation.

\textbf{Classifier.}
We use GIN \cite{DBLP:conf/iclr/XuHLJ19} as a graph classifier for evaluating the accuracy, which is one of the most popular models for graph classification and shows great performance in many domains.
The hyperparameters are searched in the same space as in their original paper \cite{DBLP:conf/iclr/XuHLJ19} to ensure that the improvement of accuracy comes from the augmentation, not from hyperparameter tuning: batch size in $\{32, 128\}$ and the dropout probability in $\{0, 0.5\}$.

\textbf{Training details.}
Following the experimental process of GIN \cite{DBLP:conf/iclr/XuHLJ19}, which we use as the classifier, we run 10-fold cross-validation for evaluation.
The indices of chosen graphs for each fold are included in the provided code repository.
The Adam optimizer \cite{DBLP:journals/corr/KingmaB14} is used, and the learning rate starts from 0.01 and decreases by half at every 50 epochs until it reaches 350 epochs.
We set the ratio $p$ of augmentation for \methodtwo, which is the only hyperparameter of our proposed approaches, to $0.4$.
All of our experiments were done at a workstation with Intel Core i7-8700 and RTX 2080.

\subsection{Accuracy of Graph Classification (Q1)}
\label{sec:exp-accuracy}

Table \ref{tab:main-accuracy} compares the accuracy of graph classification with various augmentation methods.
The Rank column is made by getting the ranks of methods in each dataset and computing their average and standard deviation over all datasets.
For example, if the rank of a method is 1, 3, and 3 in three datasets, respectively, its rank is reported as $2.3 \pm 1.2$.
The rank measures the performance of each method differently from the average accuracy: one can achieve the highest rank even though its average accuracy is low.

\methodone and \methodtwo improve the average accuracy of GIN by 1.71 and 1.75 percent points, which are 2.0$\times$ and 2.1$\times$ larger than the improvement of the best competitor, respectively.
Note that all approaches except \methodone and \methodtwo decrease the accuracy of GIN in some datasets; GraphCrop even decreases the average accuracy of GIN, implying that they give a wrong bias to the classifier by distorting the data distribution with unjustified changes.

\begin{figure}
	\begin{subfigure}{0.236\textwidth}
		\includegraphics[trim=0 2mm 0 0, clip, width=\textwidth]{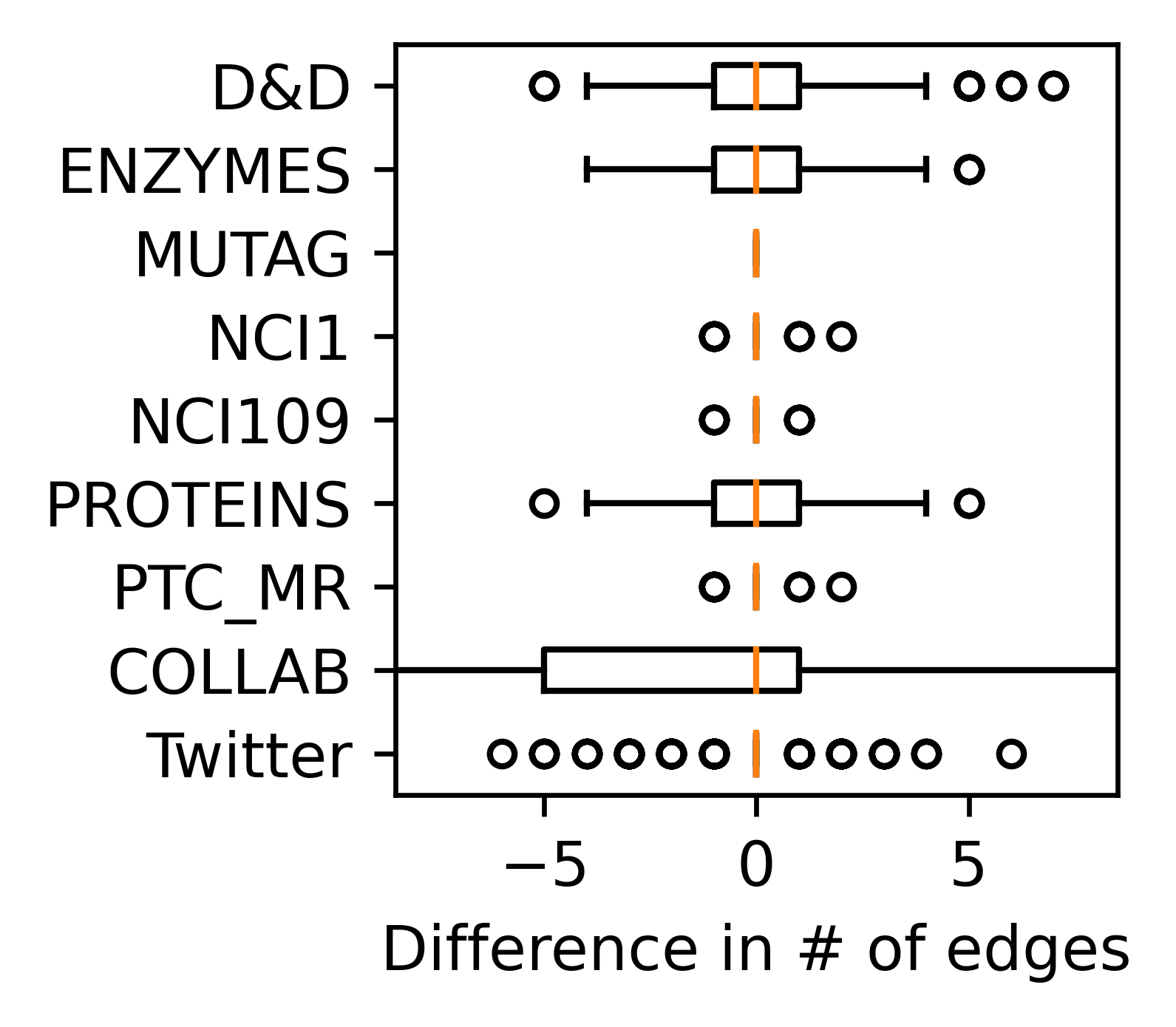}
		\caption{NodeSam}
	\end{subfigure} \hfill
	\begin{subfigure}{0.236\textwidth}
		\includegraphics[trim=0 2mm 0 0, clip, width=\textwidth]{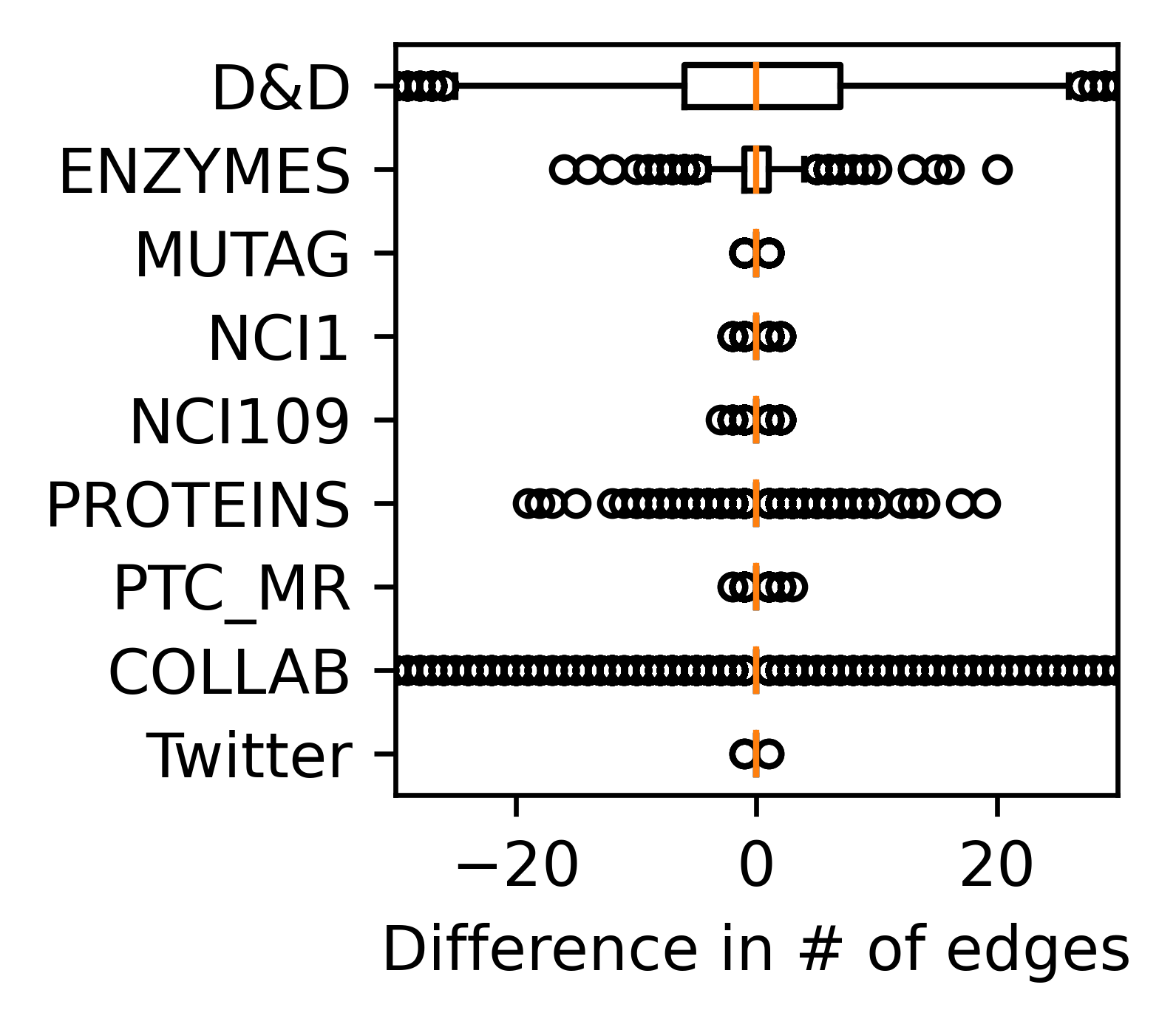}
		\caption{SubMix}
	\end{subfigure}

	\caption{
		Box plots representing the number of edges that our proposed methods change through augmentation.
		Both methods make unbiased augmentation, where the variance of changes depends on the characteristic of each dataset.
	}
	\label{fig:edge-variance}
\end{figure}

\subsection{Preserving Desired Properties (Q2)}
\label{sec:exp-properties}

The main motivation of our \methodone and \methodtwo is to satisfy the desired properties of Table \ref{table:augmentation-comparison}.
We present empirical results that support our theoretical claims given in Section \ref{sssec:nodesam-properties} and \ref{sssec:submix-properties}.

Figure \ref{fig:edge-variance} visualizes the change in the number of edges during augmentation as box plots.
Both \methodone and \methodtwo perform unbiased augmentation as shown in the orange lines that appear in the center of each plot.
On the other hand, they make a sufficient change of edges through augmentation, generating diverse examples for the training of a classifier.
\methodtwo makes a larger degree of augmentation than \methodone, especially in the D\&D dataset, as it combines multiple graphs of different structures.

\begin{figure}
	\includegraphics[width=0.46\textwidth]{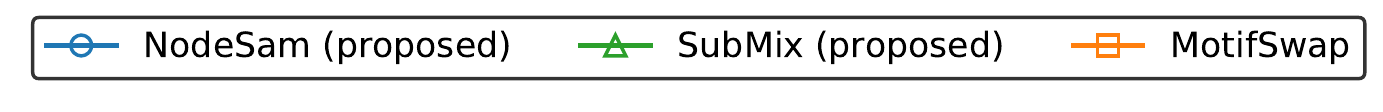}
	\vspace{-1mm}

	\begin{subfigure}{0.22\textwidth}
		\includegraphics[width=\textwidth]{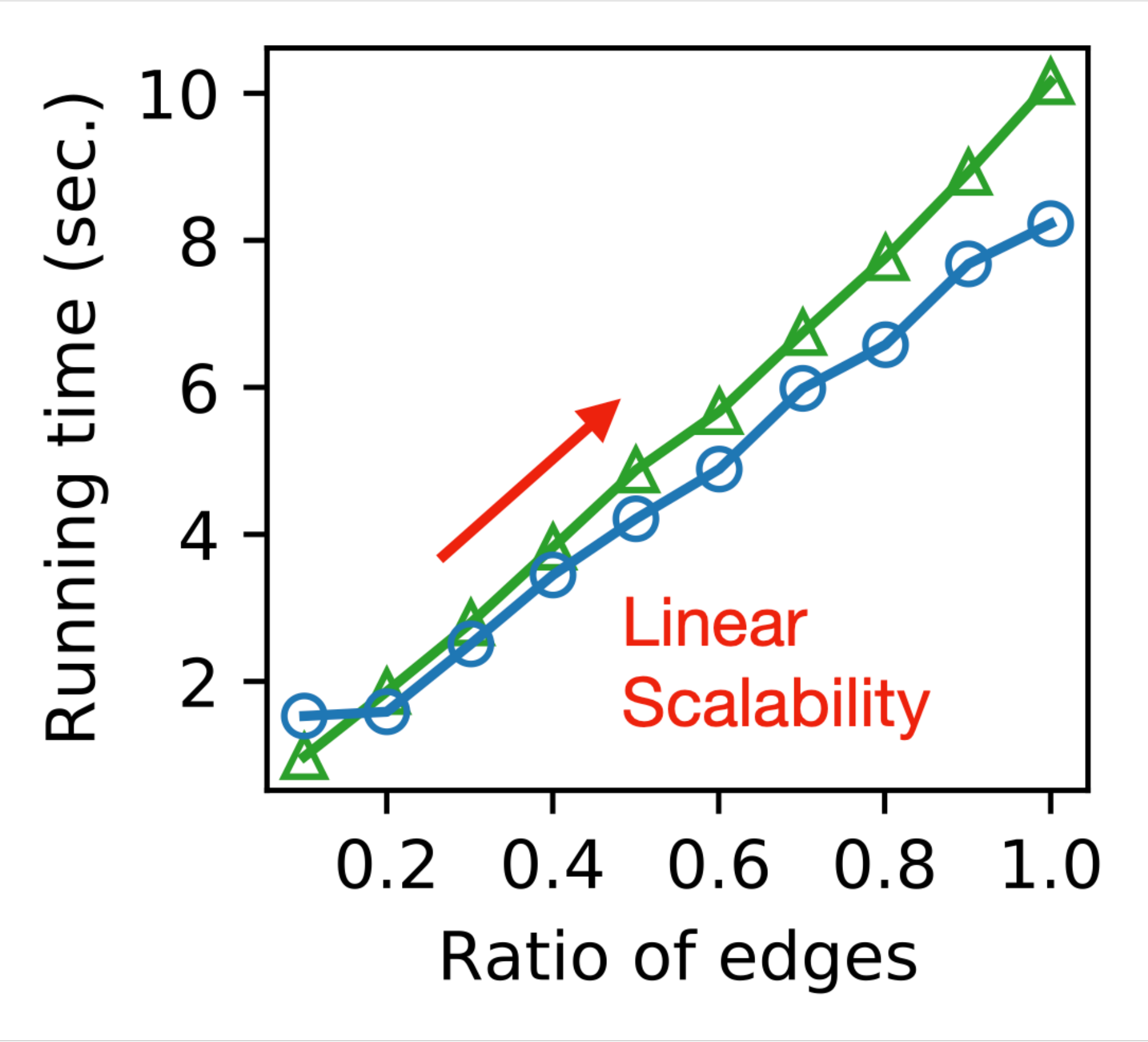}
%		\caption{NodeSam}
	\end{subfigure} \hspace{2mm}
	\begin{subfigure}{0.228\textwidth}
		\includegraphics[width=\textwidth]{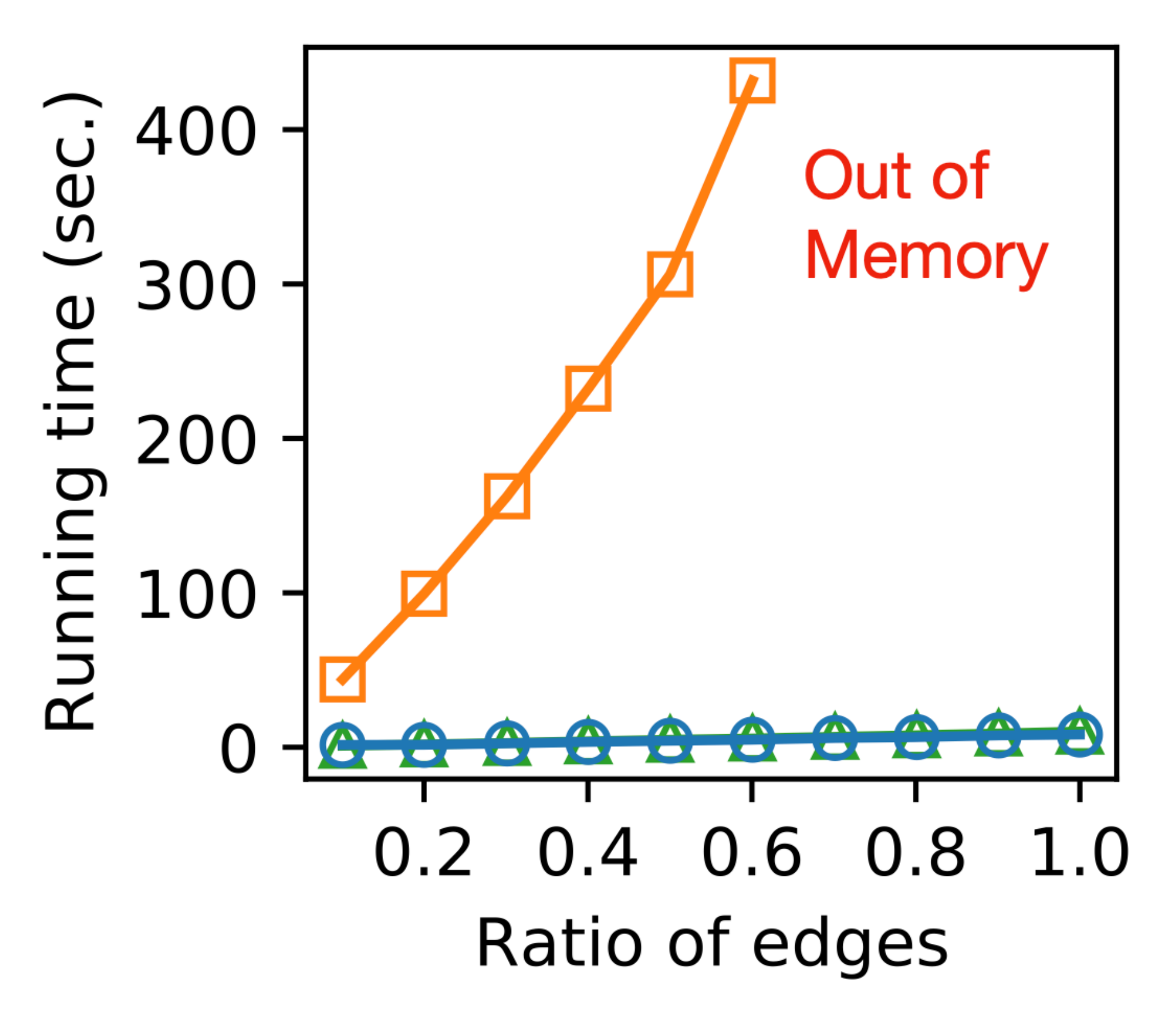}
%		\caption{SubMix}
	\end{subfigure}

	\caption{
		Computational time of our proposed methods and MotifSwap with respect to the number of edges.
		Our methods achieve linear scalability, while MotifSwap does not.
	}
	\label{fig:scalability}
\end{figure}

Figure \ref{fig:scalability} shows the scalability of \methodone and \methodtwo with the number of edges in a graph.
We use the Reddit \cite{DBLP:conf/iclr/ZengZSKP20} dataset for this experiment, which is large enough to cause scalability issues as it contains 232,965 nodes and 11,606,919 edges.
We randomly make nine subgraphs with different sizes to measure the running time of methods.
The figure shows that \methodone and \methodtwo have linear scalability with similar patterns, while the computational time of MotifSwap increases much faster, causing out-of-memory errors.
This supports our claims in Lemma \ref{lemma:complexity-1} and \ref{lemma:method-two-scalability}.

\begin{figure}
	\begin{subfigure}{0.15\textwidth}
		\includegraphics[width=\textwidth]{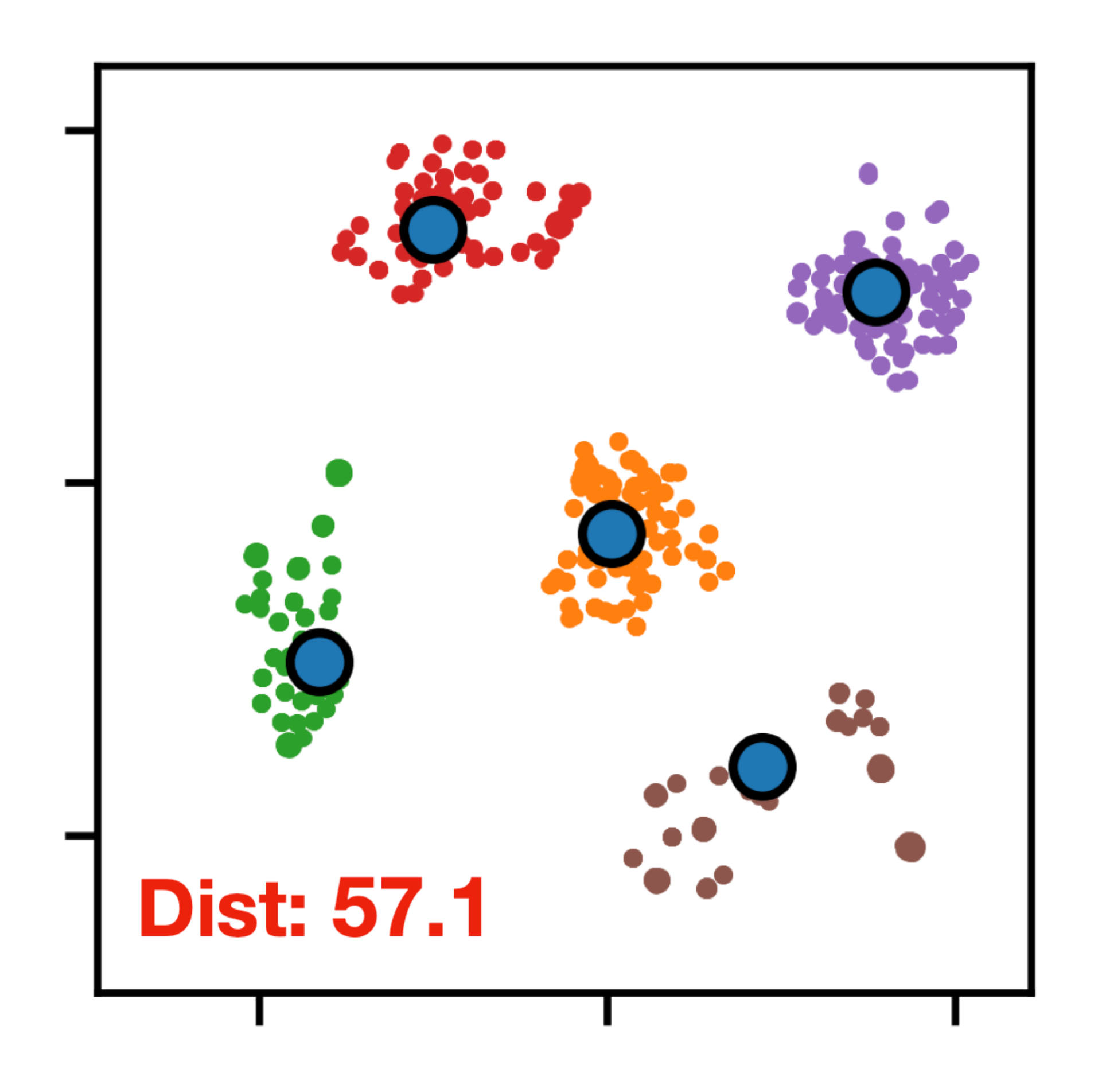}
		\caption{MotifSwap}
	\end{subfigure} \hfill
	\begin{subfigure}{0.15\textwidth}
		\includegraphics[width=\textwidth]{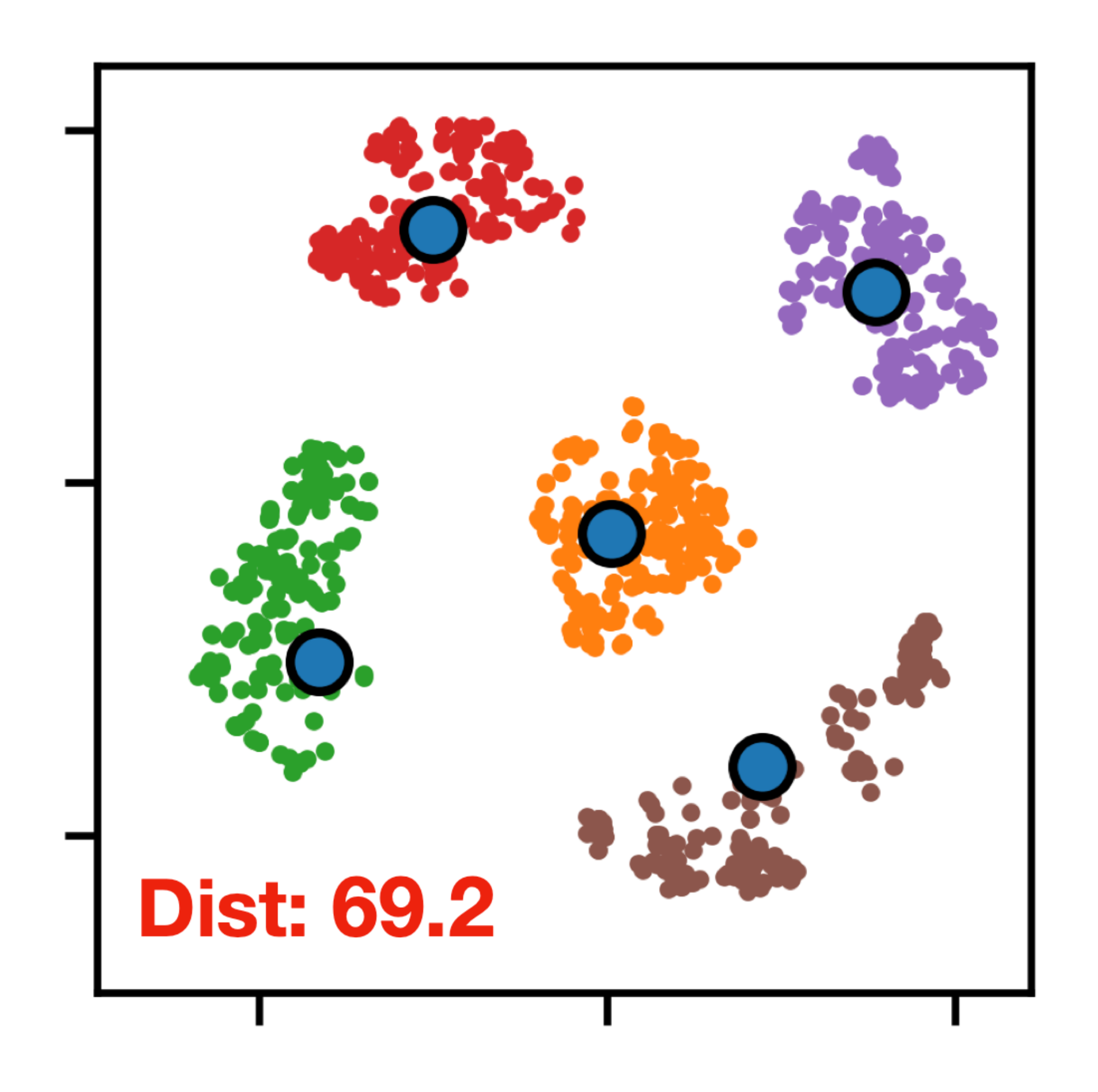}
		\caption{\methodone}
	\end{subfigure} \hfill
	\begin{subfigure}{0.15\textwidth}
		\includegraphics[width=\textwidth]{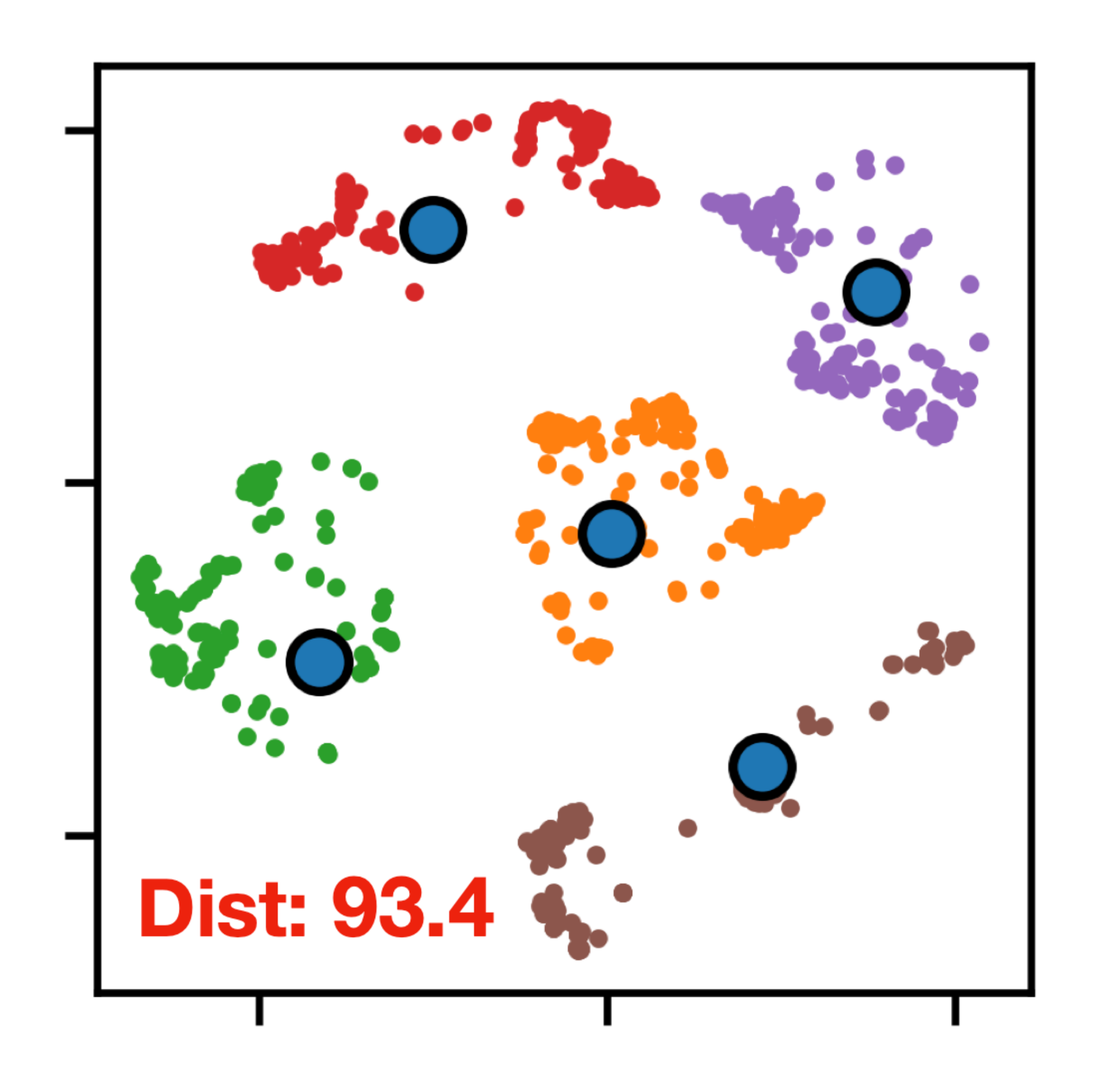}
		\caption{\methodtwo}
	\end{subfigure}

	\caption{
		Space of augmentation for various algorithms.
		The large blue points are original graphs, while the small points represent augmented graphs.
		\methodtwo has the largest space of augmentation, while MotifSwap has the smallest one.
	}
	\label{fig:aug-space}
\end{figure}

Figure \ref{fig:aug-space} visualizes the space of augmentation for our proposed algorithms and MotifSwap, the strongest baseline.
We first use the Weisfeiler-Lehman (WL) kernel \cite{DBLP:journals/jmlr/ShervashidzeSLMB11} to extract embedding vectors of graphs and then use the t-SNE algorithm \cite{van2008visualizing} to visualize them in the 2D space.
The distance value in each figure represents the average distance between each original graph and its augmented graphs in the 2D space.
\methodtwo shows the largest space of augmentation by combining multiple graphs, effectively filling in the space between given examples.
MotifSwap makes augmented graphs only near the original graphs, making the smallest average distance.

\subsection{Ablation Study (Q3)}
\label{sec:exp-ablation}

\begin{figure}
	\centering

	\begin{minipage}{0.23\textwidth}
		\centering
		\includegraphics[width=\textwidth]{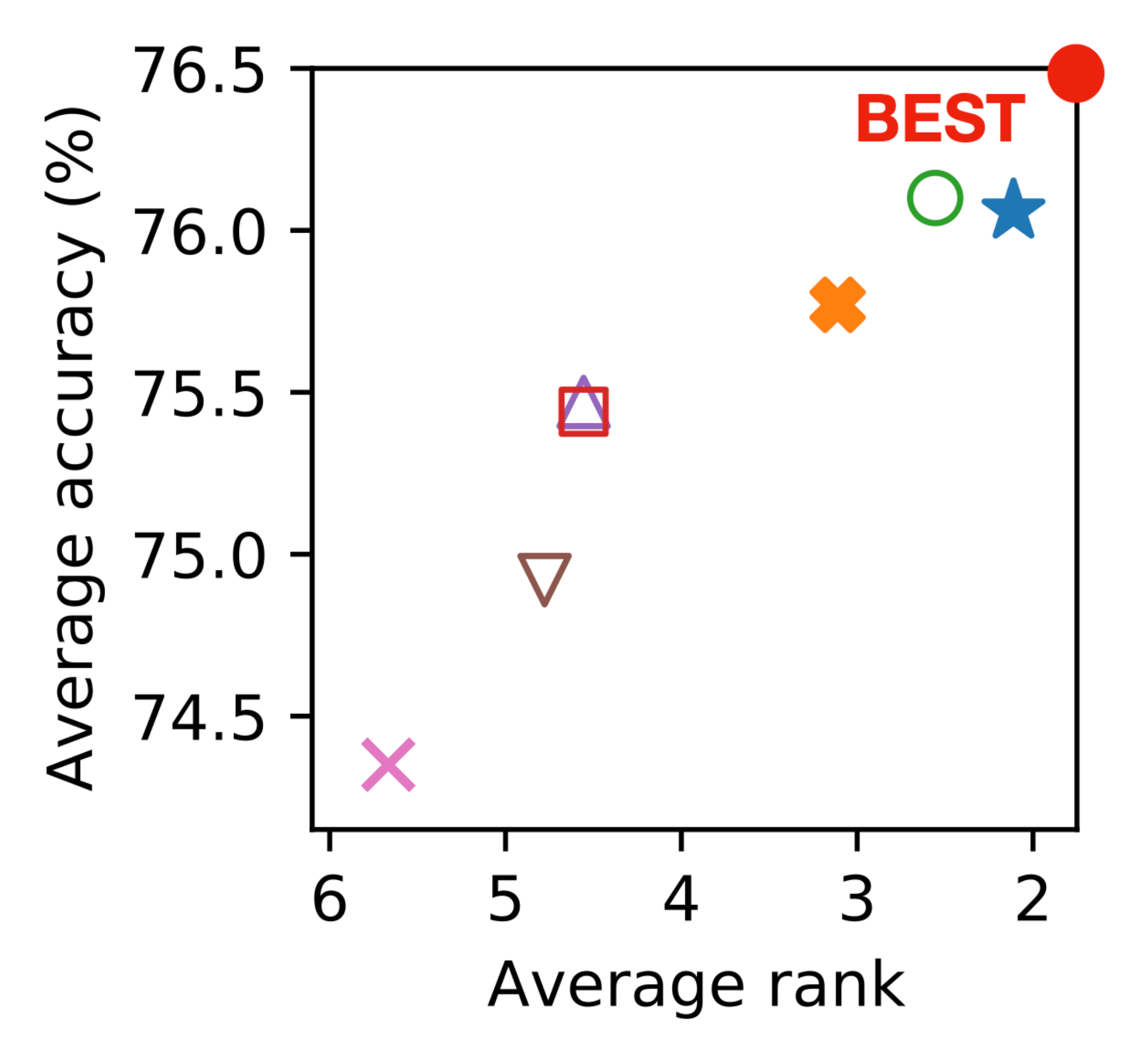}
	\end{minipage} \hspace{-2mm}
	\begin{minipage}{0.18\textwidth}
		\centering
		\vspace{-5mm}
		\includegraphics[width=\textwidth]{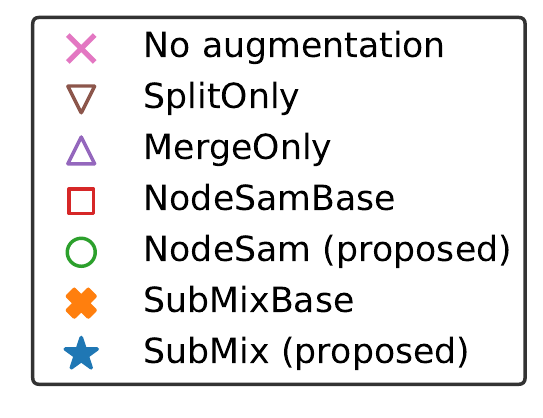}
	\end{minipage}
	
	\caption{
		Ablation study for our \methodone and \methodtwo.
		Both of them make the best average accuracy and rank based on our proposed ideas (details in Section \ref{sec:exp-ablation}).
	}
	\label{fig:ablation}
\end{figure}

We perform an ablation study for both \methodone and \methodtwo, and present the result in Figure \ref{fig:ablation}.
SplitOnly and MergeOnly refer to the split and merge operations of Algorithm \ref{alg:method-one-split} and \ref{alg:method-one-merge}, respectively.
NodeSamBase refers to the naive version of \methodone, which does not perform the adjustment operation of Algorithm \ref{alg:method-one-adjust}.
SubMixBase refers to the naive version of \methodtwo, which selects the target sets of nodes for the replacement uniformly at random, without using the diffusion operation of Algorithm  \ref{alg:method-two-sample}.
The vertical and horizontal axes of the figure are the average accuracy and rank over all datasets, corresponding to the last two columns of Table \ref{tab:main-accuracy}, respectively.

We have two observations from Figure \ref{fig:ablation}.
First, \methodone and \methodtwo with all techniques achieve the best accuracy among all versions, showing that our proposed techniques are effective for improving their performance by satisfying the desired properties.
Second, every basic version of our approaches still improves the baseline significantly, which is to use raw graphs without augmentation.
%This demonstrates the effectiveness of our basic ideas, which work as building blocks of \methodone and \methodtwo.
This is clear when compared with existing methods shown in Table \ref{tab:main-accuracy} that make marginal improvements or even decrease the baseline accuracy.
Such improvement exhibits the effectiveness of our ideas, which are building blocks of \methodone and \methodtwo.

We also perform an additional study for \methodone and its basic versions in Appendix \ref{appendix:ablation}, which shows that NodeSamBase without the adjustment operation is likely to decrease the number of edges due to the merge operation that eliminates additional triangles.

\section{Related Works}
\label{sec:related-works}

We review related works for graph augmentation, graph classification, and image augmentation.

\textbf{Model-agnostic graph augmentation.}
Our work focuses on model-agnostic augmentation, which is to perform augmentation independently from target models.
Previous works can be categorized by the purpose of augmentation: node-level tasks \cite{DBLP:conf/kdd/WangWLCLH20, DBLP:conf/iclr/RongHXH20, DBLP:conf/wsdm/YooKSCZ20}, graph-level tasks \cite{DBLP:journals/corr/abs-2009-10564, DBLP:journals/jvis/FuMWLZZSL21, DBLP:conf/cikm/ZhouSX20}, or graph contrastive learning \cite{DBLP:conf/icml/YouCSW21, DBLP:conf/nips/YouCSCWS20}.
%Some approaches utilize generative models to generate plausible graphs rather than changing existing ones \cite{DBLP:conf/aaai/WangWWZZZXG18, DBLP:journals/fdata/ZhouZXH19}.
Such methods rely on heuristic operations that make no theoretical guarantee for the degree of augmentation or the preservation of graph properties.
We propose five desired properties for effective graph augmentation and show that our approaches satisfy all of them, resulting in the best accuracy in classification.

\begin{comment}
Model-agnostic methods work independently from the target model and can be used generally in various settings.
NodeAug \cite{DBLP:conf/kdd/WangWLCLH20} augments the feature and local neighborhood of a target node for node-level augmentation.
GraphCrop \cite{DBLP:journals/corr/abs-2009-10564} and TS-Extractor \cite{DBLP:journals/jvis/FuMWLZZSL21} extract subgraphs for graph-level augmentation.
DropEdge \cite{DBLP:conf/iclr/RongHXH20} removes a portion of edges for better training of graph neural networks.
Song et al. \cite{DBLP:journals/corr/abs-2104-02478} enhance node features using topology information, while You et al. \cite{DBLP:conf/nips/YouCSCWS20} augment graphs for graph contrastive learning.
Zhou et al. \cite{DBLP:conf/cikm/ZhouSX20} propose an augmentation approach that swaps the edges in open triangles to preserve the properties of the original graph.
\end{comment}

%GraphGAN \cite{DBLP:conf/aaai/WangWWZZZXG18} and Misc-GAN \cite{DBLP:journals/fdata/ZhouZXH19} utilize generative models to create more plausible graphs for augmentation.

\textbf{Model-specific graph augmentation.}
Model-specific methods for graph augmentation make changes optimized for the target model.
For example, changing the representation vectors of graphs learned by the target classifier by a mixup algorithm \cite{DBLP:conf/www/WangWLCH21} is model-specific. 
Such model-specific approaches include manifold mixup \cite{DBLP:journals/corr/abs-1909-11715}, dynamic augmentation of edges \cite{DBLP:journals/corr/abs-2006-06830, DBLP:conf/aaai/ChenLLLZS20}, and adversarial perturbations \cite{DBLP:journals/corr/abs-1902-08226, DBLP:journals/corr/abs-2010-09891}.
Methods for graph adversarial training \cite{DBLP:conf/icml/DaiLTHWZS18, DBLP:conf/iclr/ZugnerG19, DBLP:conf/ijcai/XuC0CWHL19, DBLP:conf/nips/0001DM20} also belong to this category as they require the predictions of target models.
Such approaches participate directly in the training process, and thus cannot be used generally in various settings.

%\begin{table}
%	\centering
%	\caption{\red{%
%		Ablation study of \methodone and \methodtwo by the average accuracy on all datasets.
%		Our ideas improve the accuracy of both models.
%	}}
%	\begin{tabular}{l|c}
%		\toprule
%		\textbf{Method} & \textbf{Accuracy} \\
%		\midrule
%		Baseline & 74.42 $\pm$ 3.14 \\
%		\midrule
%		Split only & 75.18 $\pm$ 2.77 \\
%		Merge only & 75.46 $\pm$ 3.03 \\
%		\methodone-base & 76.28 $\pm$ 3.48 \\
%		\methodone & 76.78 $\pm$ 2.94 \\
%		\midrule
%		\methodtwo-base & 75.84 $\pm$ 3.16\\
%		\methodtwo & 76.22 $\pm$ 2.69 \\
%		\bottomrule
%	\end{tabular}
%	\label{table:ablation}
%\end{table}

\textbf{Graph classification.}
Graph classification is a core problem in the graph domain, which is to predict the label of each graph.
Compared to node classification \cite{DBLP:conf/iclr/KipfW17} that focuses on the properties of individual nodes, graph classification aims at extracting meaningful substructures of nodes to make a better representation of a graph.
A traditional way to solve the problem is to utilize kernel methods \cite{DBLP:journals/jmlr/ShervashidzeSLMB11, DBLP:conf/kdd/YanardagV15, DBLP:journals/corr/NarayananCCLS16, DBLP:conf/icml/RieckBB19}, while graph neural networks \cite{DBLP:conf/iclr/KipfW17, DBLP:conf/iclr/XuHLJ19, DBLP:conf/nips/KlicperaWG19, DBLP:conf/ijcai/YooJK19, DBLP:conf/icdm/YooKYKJK21, DBLP:journals/corr/abs-2012-14191} have shown a better performance with advanced pooling methods \cite{DBLP:conf/nips/YingY0RHL18, DBLP:conf/kdd/0001WAT19, DBLP:conf/icml/BianchiGA20} to aggregate node embeddings.
We use graph classification as a downstream task for evaluating augmented graphs, since accurate classification requires separating the core information of each graph from random and unimportant components.

%Devising an augmentation algorithm for accurate graph classification requires separating the core information that determines the characteristic of each graph from the random and unimportant components.

%On the other hand, a simple heuristic augmentation can significantly improve the accuracy in node-level tasks such as node classification \cite{DBLP:conf/kdd/WangWLCLH20, DBLP:conf/iclr/RongHXH20}.
%We focus on a challenging scenario that requires a careful design of augmentation algorithms for preserving the core properties of graphs.
%We use GIN \cite{DBLP:conf/iclr/XuHLJ19} to evaluate the effect of augmentation for graph classification, which is popular and still considered as one of the state-of-the-art models for the problem.

\textbf{Image augmentation.}
Data augmentation is studied actively in the image domain.
Basic approaches include color and geometric transformations \cite{DBLP:journals/corr/GaldranAMSAGACM17, DBLP:journals/pami/RoccoAS19}, masking \cite{DBLP:journals/corr/abs-1708-04552, DBLP:conf/aaai/Zhong0KL020, DBLP:conf/iccv/SinghL17}, adversarial perturbations \cite{DBLP:conf/cvpr/Moosavi-Dezfooli16}, pixel-based mixing \cite{DBLP:conf/iclr/ZhangCDL18, DBLP:conf/cvpr/LeeZAL20}, and generative models \cite{DBLP:journals/ijon/Frid-AdarDKAGG18, DBLP:conf/pakdd/ZhuLLWQ18}.
Such approaches utilize the characteristic of images where slight changes of pixel values do not change the semantic information of each image.
On the other hand, patch-based mixing approaches that replace a random patch of an image to that of another image \cite{DBLP:conf/iccv/YunHCOYC19, DBLP:conf/icml/KimCS20, DBLP:journals/tcsv/TakahashiMU20} give us a motivation to propose \methodtwo, which replaces an induced subgraph of a graph, instead of an image patch.
This is based on the idea that images are grid-structured graphs whose pixels correspond to individual nodes.

\section{Conclusion} \label{sec:conclusion}

We propose \methodone (\methodonelong) and \methodtwo (\methodtwolong), effective algorithms for model-agnostic graph augmentation.
We first present desired properties for graph augmentation including the preservation of original properties, the guarantee of sufficient augmentation, and the linear scalability.
Then, we show that both \methodone and \methodtwo achieve all properties unlike all of the existing approaches.
\methodone aims to make a minimal change of the graph structure by performing opposite operations at once: splitting a node and merging two nodes.
On the other hand, \methodtwo aims to increase the degree of augmentation by combining random subgraphs of different graphs with different labels.
Experiments on nine datasets show that \methodone and \methodtwo achieve the best accuracy in graph classification with up to 2.1$\times$ larger improvement of accuracy compared with previous approaches.

\begin{acks}
{ \small
This work was supported by Institute of Information \& communications Technology Planning \& Evaluation(IITP) grant funded by the Korea government(MSIT) [No.2020-0-00894, Flexible and Efficient Model Compression Method for Various Applications and Environments], 
[No.2021-0-01343, Artificial Intelligence Graduate School Program (Seoul National University)],
and 
[NO.2021-0-0268, Artificial Intelligence Innovation Hub (Artificial Intelligence Institute, Seoul National University)]. 
The Institute of Engineering Research at Seoul National University provided research facilities for this work.
The ICT at Seoul National University provides research facilities for this study.
U Kang is the corresponding author.
}
\end{acks}

\clearpage

\bibliographystyle{ACM-Reference-Format}
\bibliography{paper}


\begin{thebibliography}{57}

%%% the Web address.

\ifx \showCODEN    \undefined \def \showCODEN     #1{\unskip}     \fi
\ifx \showDOI      \undefined \def \showDOI       #1{#1}\fi
\ifx \showISBNx    \undefined \def \showISBNx     #1{\unskip}     \fi
\ifx \showISBNxiii \undefined \def \showISBNxiii  #1{\unskip}     \fi
\ifx \showISSN     \undefined \def \showISSN      #1{\unskip}     \fi
\ifx \showLCCN     \undefined \def \showLCCN      #1{\unskip}     \fi
\ifx \shownote     \undefined \def \shownote      #1{#1}          \fi
\ifx \showarticletitle \undefined \def \showarticletitle #1{#1}   \fi
\ifx \showURL      \undefined \def \showURL       {\relax}        \fi
% invisible to TeX
\providecommand\bibfield[2]{#2}
\providecommand\bibinfo[2]{#2}
\providecommand\natexlab[1]{#1}
\providecommand\showeprint[2][]{arXiv:#2}

\bibitem[\protect\citeauthoryear{Bianchi, Grattarola, and Alippi}{Bianchi
  et~al\mbox{.}}{2020}]%
        {DBLP:conf/icml/BianchiGA20}
\bibfield{author}{\bibinfo{person}{Filippo~Maria Bianchi},
  \bibinfo{person}{Daniele Grattarola}, {and} \bibinfo{person}{Cesare Alippi}.}
  \bibinfo{year}{2020}\natexlab{}.
\newblock \showarticletitle{Spectral Clustering with Graph Neural Networks for
  Graph Pooling}. In \bibinfo{booktitle}{\emph{ICML}}.
\newblock


\bibitem[\protect\citeauthoryear{Book}{Book}{1974}]%
        {DBLP:journals/jcss/Book74}
\bibfield{author}{\bibinfo{person}{Ronald~V. Book}.}
  \bibinfo{year}{1974}\natexlab{}.
\newblock \showarticletitle{Comparing Complexity Classes}.
\newblock \bibinfo{journal}{\emph{J. Comput. Syst. Sci.}} \bibinfo{volume}{9},
  \bibinfo{number}{2} (\bibinfo{year}{1974}), \bibinfo{pages}{213--229}.
\newblock
\urldef\tempurl%
\url{https://doi.org/10.1016/S0022-0000(74)80008-5}
\showDOI{\tempurl}


\bibitem[\protect\citeauthoryear{Chen, Lin, Li, Li, Zhou, and Sun}{Chen
  et~al\mbox{.}}{2020}]%
        {DBLP:conf/aaai/ChenLLLZS20}
\bibfield{author}{\bibinfo{person}{Deli Chen}, \bibinfo{person}{Yankai Lin},
  \bibinfo{person}{Wei Li}, \bibinfo{person}{Peng Li}, \bibinfo{person}{Jie
  Zhou}, {and} \bibinfo{person}{Xu Sun}.} \bibinfo{year}{2020}\natexlab{}.
\newblock \showarticletitle{Measuring and Relieving the Over-Smoothing Problem
  for Graph Neural Networks from the Topological View}. In
  \bibinfo{booktitle}{\emph{AAAI}}.
\newblock


\bibitem[\protect\citeauthoryear{Dai, Li, Tian, Huang, Wang, Zhu, and Song}{Dai
  et~al\mbox{.}}{2018}]%
        {DBLP:conf/icml/DaiLTHWZS18}
\bibfield{author}{\bibinfo{person}{Hanjun Dai}, \bibinfo{person}{Hui Li},
  \bibinfo{person}{Tian Tian}, \bibinfo{person}{Xin Huang},
  \bibinfo{person}{Lin Wang}, \bibinfo{person}{Jun Zhu}, {and}
  \bibinfo{person}{Le Song}.} \bibinfo{year}{2018}\natexlab{}.
\newblock \showarticletitle{Adversarial Attack on Graph Structured Data}. In
  \bibinfo{booktitle}{\emph{ICML}}.
\newblock


\bibitem[\protect\citeauthoryear{Devries and Taylor}{Devries and
  Taylor}{2017}]%
        {DBLP:journals/corr/abs-1708-04552}
\bibfield{author}{\bibinfo{person}{Terrance Devries} {and}
  \bibinfo{person}{Graham~W. Taylor}.} \bibinfo{year}{2017}\natexlab{}.
\newblock \showarticletitle{Improved Regularization of Convolutional Neural
  Networks with Cutout}.
\newblock \bibinfo{journal}{\emph{CoRR}}  \bibinfo{volume}{abs/1708.04552}
  (\bibinfo{year}{2017}).
\newblock
\showeprint[arxiv]{1708.04552}


\bibitem[\protect\citeauthoryear{Du and Tong}{Du and Tong}{2019}]%
        {DBLP:conf/cikm/DuT19}
\bibfield{author}{\bibinfo{person}{Boxin Du} {and} \bibinfo{person}{Hanghang
  Tong}.} \bibinfo{year}{2019}\natexlab{}.
\newblock \showarticletitle{MrMine: Multi-resolution Multi-network Embedding}.
  In \bibinfo{booktitle}{\emph{CIKM}}.
\newblock


\bibitem[\protect\citeauthoryear{Feng, He, Tang, and Chua}{Feng
  et~al\mbox{.}}{2019}]%
        {DBLP:journals/corr/abs-1902-08226}
\bibfield{author}{\bibinfo{person}{Fuli Feng}, \bibinfo{person}{Xiangnan He},
  \bibinfo{person}{Jie Tang}, {and} \bibinfo{person}{Tat{-}Seng Chua}.}
  \bibinfo{year}{2019}\natexlab{}.
\newblock \showarticletitle{Graph Adversarial Training: Dynamically
  Regularizing Based on Graph Structure}.
\newblock \bibinfo{journal}{\emph{CoRR}}  \bibinfo{volume}{abs/1902.08226}
  (\bibinfo{year}{2019}).
\newblock
\showeprint[arxiv]{1902.08226}


\bibitem[\protect\citeauthoryear{Feng, Gangal, Wei, Chandar, Vosoughi,
  Mitamura, and Hovy}{Feng et~al\mbox{.}}{2021}]%
        {DBLP:conf/acl/FengGWCVMH21}
\bibfield{author}{\bibinfo{person}{Steven~Y. Feng}, \bibinfo{person}{Varun
  Gangal}, \bibinfo{person}{Jason Wei}, \bibinfo{person}{Sarath Chandar},
  \bibinfo{person}{Soroush Vosoughi}, \bibinfo{person}{Teruko Mitamura}, {and}
  \bibinfo{person}{Eduard~H. Hovy}.} \bibinfo{year}{2021}\natexlab{}.
\newblock \showarticletitle{A Survey of Data Augmentation Approaches for
  {NLP}}. In \bibinfo{booktitle}{\emph{Findings of ACL}}.
\newblock


\bibitem[\protect\citeauthoryear{Frid{-}Adar, Diamant, Klang, Amitai,
  Goldberger, and Greenspan}{Frid{-}Adar et~al\mbox{.}}{2018}]%
        {DBLP:journals/ijon/Frid-AdarDKAGG18}
\bibfield{author}{\bibinfo{person}{Maayan Frid{-}Adar}, \bibinfo{person}{Idit
  Diamant}, \bibinfo{person}{Eyal Klang}, \bibinfo{person}{Michal Amitai},
  \bibinfo{person}{Jacob Goldberger}, {and} \bibinfo{person}{Hayit Greenspan}.}
  \bibinfo{year}{2018}\natexlab{}.
\newblock \showarticletitle{GAN-based synthetic medical image augmentation for
  increased {CNN} performance in liver lesion classification}.
\newblock \bibinfo{journal}{\emph{Neurocomputing}}  \bibinfo{volume}{321}
  (\bibinfo{year}{2018}), \bibinfo{pages}{321--331}.
\newblock


\bibitem[\protect\citeauthoryear{Fu, Mao, Wang, Lin, Zhang, Zhan, Sun, and
  Li}{Fu et~al\mbox{.}}{2021}]%
        {DBLP:journals/jvis/FuMWLZZSL21}
\bibfield{author}{\bibinfo{person}{Kun Fu}, \bibinfo{person}{Tingyun Mao},
  \bibinfo{person}{Yang Wang}, \bibinfo{person}{Daoyu Lin},
  \bibinfo{person}{Yuanben Zhang}, \bibinfo{person}{Junjian Zhan},
  \bibinfo{person}{Xian Sun}, {and} \bibinfo{person}{Feng Li}.}
  \bibinfo{year}{2021}\natexlab{}.
\newblock \showarticletitle{TS-Extractor: large graph exploration via subgraph
  extraction based on topological and semantic information}.
\newblock \bibinfo{journal}{\emph{J. Vis.}} \bibinfo{volume}{24},
  \bibinfo{number}{1} (\bibinfo{year}{2021}), \bibinfo{pages}{173--190}.
\newblock


\bibitem[\protect\citeauthoryear{Galdran, Alvarez{-}Gila, Meyer, Saratxaga,
  Araujo, Garrote, Aresta, Costa, Mendon{\c{c}}a, and Campilho}{Galdran
  et~al\mbox{.}}{2017}]%
        {DBLP:journals/corr/GaldranAMSAGACM17}
\bibfield{author}{\bibinfo{person}{Adrian Galdran}, \bibinfo{person}{Aitor
  Alvarez{-}Gila}, \bibinfo{person}{Maria~In{\^{e}}s Meyer},
  \bibinfo{person}{Cristina~L{\'{o}}pez Saratxaga}, \bibinfo{person}{Teresa
  Araujo}, \bibinfo{person}{Est{\'{\i}}baliz Garrote},
  \bibinfo{person}{Guilherme Aresta}, \bibinfo{person}{Pedro Costa},
  \bibinfo{person}{Ana~Maria Mendon{\c{c}}a}, {and}
  \bibinfo{person}{Aur{\'{e}}lio J.~C. Campilho}.}
  \bibinfo{year}{2017}\natexlab{}.
\newblock \showarticletitle{Data-Driven Color Augmentation Techniques for Deep
  Skin Image Analysis}.
\newblock \bibinfo{journal}{\emph{CoRR}}  \bibinfo{volume}{abs/1703.03702}
  (\bibinfo{year}{2017}).
\newblock
\showeprint[arxiv]{1703.03702}


\bibitem[\protect\citeauthoryear{Jo, Yoo, and Kang}{Jo et~al\mbox{.}}{2018}]%
        {DBLP:conf/wsdm/JoYK18}
\bibfield{author}{\bibinfo{person}{Saehan Jo}, \bibinfo{person}{Jaemin Yoo},
  {and} \bibinfo{person}{U Kang}.} \bibinfo{year}{2018}\natexlab{}.
\newblock \showarticletitle{Fast and Scalable Distributed Loopy Belief
  Propagation on Real-World Graphs}. In \bibinfo{booktitle}{\emph{WSDM}}.
\newblock


\bibitem[\protect\citeauthoryear{Jung, Yoo, and Kang}{Jung
  et~al\mbox{.}}{2020}]%
        {DBLP:journals/corr/abs-2012-14191}
\bibfield{author}{\bibinfo{person}{Jinhong Jung}, \bibinfo{person}{Jaemin Yoo},
  {and} \bibinfo{person}{U Kang}.} \bibinfo{year}{2020}\natexlab{}.
\newblock \showarticletitle{Signed Graph Diffusion Network}.
\newblock \bibinfo{journal}{\emph{CoRR}}  \bibinfo{volume}{abs/2012.14191}
  (\bibinfo{year}{2020}).
\newblock
\showeprint[arXiv]{2012.14191}


\bibitem[\protect\citeauthoryear{Kim, Choo, and Song}{Kim
  et~al\mbox{.}}{2020}]%
        {DBLP:conf/icml/KimCS20}
\bibfield{author}{\bibinfo{person}{Jang{-}Hyun Kim}, \bibinfo{person}{Wonho
  Choo}, {and} \bibinfo{person}{Hyun~Oh Song}.}
  \bibinfo{year}{2020}\natexlab{}.
\newblock \showarticletitle{Puzzle Mix: Exploiting Saliency and Local
  Statistics for Optimal Mixup}. In \bibinfo{booktitle}{\emph{ICML}}.
\newblock


\bibitem[\protect\citeauthoryear{Kingma and Ba}{Kingma and Ba}{2015}]%
        {DBLP:journals/corr/KingmaB14}
\bibfield{author}{\bibinfo{person}{Diederik~P. Kingma} {and}
  \bibinfo{person}{Jimmy Ba}.} \bibinfo{year}{2015}\natexlab{}.
\newblock \showarticletitle{Adam: {A} Method for Stochastic Optimization}. In
  \bibinfo{booktitle}{\emph{ICLR}}.
\newblock


\bibitem[\protect\citeauthoryear{Kipf and Welling}{Kipf and Welling}{2017}]%
        {DBLP:conf/iclr/KipfW17}
\bibfield{author}{\bibinfo{person}{Thomas~N. Kipf} {and} \bibinfo{person}{Max
  Welling}.} \bibinfo{year}{2017}\natexlab{}.
\newblock \showarticletitle{Semi-Supervised Classification with Graph
  Convolutional Networks}. In \bibinfo{booktitle}{\emph{ICLR}}.
\newblock


\bibitem[\protect\citeauthoryear{Klicpera, Wei{\ss}enberger, and
  G{\"{u}}nnemann}{Klicpera et~al\mbox{.}}{2019}]%
        {DBLP:conf/nips/KlicperaWG19}
\bibfield{author}{\bibinfo{person}{Johannes Klicpera}, \bibinfo{person}{Stefan
  Wei{\ss}enberger}, {and} \bibinfo{person}{Stephan G{\"{u}}nnemann}.}
  \bibinfo{year}{2019}\natexlab{}.
\newblock \showarticletitle{Diffusion Improves Graph Learning}. In
  \bibinfo{booktitle}{\emph{NeurIPS}}.
\newblock


\bibitem[\protect\citeauthoryear{Kong, Li, Ding, Wu, Zhu, Ghanem, Taylor, and
  Goldstein}{Kong et~al\mbox{.}}{2020}]%
        {DBLP:journals/corr/abs-2010-09891}
\bibfield{author}{\bibinfo{person}{Kezhi Kong}, \bibinfo{person}{Guohao Li},
  \bibinfo{person}{Mucong Ding}, \bibinfo{person}{Zuxuan Wu},
  \bibinfo{person}{Chen Zhu}, \bibinfo{person}{Bernard Ghanem},
  \bibinfo{person}{Gavin Taylor}, {and} \bibinfo{person}{Tom Goldstein}.}
  \bibinfo{year}{2020}\natexlab{}.
\newblock \showarticletitle{{FLAG:} Adversarial Data Augmentation for Graph
  Neural Networks}.
\newblock \bibinfo{journal}{\emph{CoRR}}  \bibinfo{volume}{abs/2010.09891}
  (\bibinfo{year}{2020}).
\newblock
\showeprint[arxiv]{2010.09891}


\bibitem[\protect\citeauthoryear{Lee, Zaheer, Astrid, and Lee}{Lee
  et~al\mbox{.}}{2020}]%
        {DBLP:conf/cvpr/LeeZAL20}
\bibfield{author}{\bibinfo{person}{Jin{-}Ha Lee},
  \bibinfo{person}{Muhammad~Zaigham Zaheer}, \bibinfo{person}{Marcella Astrid},
  {and} \bibinfo{person}{Seung{-}Ik Lee}.} \bibinfo{year}{2020}\natexlab{}.
\newblock \showarticletitle{SmoothMix: a Simple Yet Effective Data Augmentation
  to Train Robust Classifiers}. In \bibinfo{booktitle}{\emph{CVPR}}.
\newblock


\bibitem[\protect\citeauthoryear{Ma, Ding, and Mei}{Ma et~al\mbox{.}}{2020}]%
        {DBLP:conf/nips/0001DM20}
\bibfield{author}{\bibinfo{person}{Jiaqi Ma}, \bibinfo{person}{Shuangrui Ding},
  {and} \bibinfo{person}{Qiaozhu Mei}.} \bibinfo{year}{2020}\natexlab{}.
\newblock \showarticletitle{Towards More Practical Adversarial Attacks on Graph
  Neural Networks}. In \bibinfo{booktitle}{\emph{NeurIPS}}.
\newblock


\bibitem[\protect\citeauthoryear{Ma, Wang, Aggarwal, and Tang}{Ma
  et~al\mbox{.}}{2019}]%
        {DBLP:conf/kdd/0001WAT19}
\bibfield{author}{\bibinfo{person}{Yao Ma}, \bibinfo{person}{Suhang Wang},
  \bibinfo{person}{Charu~C. Aggarwal}, {and} \bibinfo{person}{Jiliang Tang}.}
  \bibinfo{year}{2019}\natexlab{}.
\newblock \showarticletitle{Graph Convolutional Networks with EigenPooling}. In
  \bibinfo{booktitle}{\emph{KDD}}.
\newblock


\bibitem[\protect\citeauthoryear{Moosavi{-}Dezfooli, Fawzi, and
  Frossard}{Moosavi{-}Dezfooli et~al\mbox{.}}{2016}]%
        {DBLP:conf/cvpr/Moosavi-Dezfooli16}
\bibfield{author}{\bibinfo{person}{Seyed{-}Mohsen Moosavi{-}Dezfooli},
  \bibinfo{person}{Alhussein Fawzi}, {and} \bibinfo{person}{Pascal Frossard}.}
  \bibinfo{year}{2016}\natexlab{}.
\newblock \showarticletitle{DeepFool: {A} Simple and Accurate Method to Fool
  Deep Neural Networks}. In \bibinfo{booktitle}{\emph{CVPR}}.
\newblock


\bibitem[\protect\citeauthoryear{Morris, Kriege, Bause, Kersting, Mutzel, and
  Neumann}{Morris et~al\mbox{.}}{2020}]%
        {DBLP:journals/corr/abs-2007-08663}
\bibfield{author}{\bibinfo{person}{Christopher Morris},
  \bibinfo{person}{Nils~M. Kriege}, \bibinfo{person}{Franka Bause},
  \bibinfo{person}{Kristian Kersting}, \bibinfo{person}{Petra Mutzel}, {and}
  \bibinfo{person}{Marion Neumann}.} \bibinfo{year}{2020}\natexlab{}.
\newblock \showarticletitle{TUDataset: {A} collection of benchmark datasets for
  learning with graphs}.
\newblock \bibinfo{journal}{\emph{CoRR}}  \bibinfo{volume}{abs/2007.08663}
  (\bibinfo{year}{2020}).
\newblock
\showeprint[arxiv]{2007.08663}


\bibitem[\protect\citeauthoryear{Narayanan, Chandramohan, Chen, Liu, and
  Saminathan}{Narayanan et~al\mbox{.}}{2016}]%
        {DBLP:journals/corr/NarayananCCLS16}
\bibfield{author}{\bibinfo{person}{Annamalai Narayanan},
  \bibinfo{person}{Mahinthan Chandramohan}, \bibinfo{person}{Lihui Chen},
  \bibinfo{person}{Yang Liu}, {and} \bibinfo{person}{Santhoshkumar
  Saminathan}.} \bibinfo{year}{2016}\natexlab{}.
\newblock \showarticletitle{subgraph2vec: Learning Distributed Representations
  of Rooted Sub-graphs from Large Graphs}.
\newblock \bibinfo{journal}{\emph{CoRR}}  \bibinfo{volume}{abs/1606.08928}
  (\bibinfo{year}{2016}).
\newblock
\showeprint[arxiv]{1606.08928}


\bibitem[\protect\citeauthoryear{Pan, Wu, and Zhu}{Pan et~al\mbox{.}}{2015}]%
        {DBLP:journals/tkde/PanWZ15}
\bibfield{author}{\bibinfo{person}{Shirui Pan}, \bibinfo{person}{Jia Wu}, {and}
  \bibinfo{person}{Xingquan Zhu}.} \bibinfo{year}{2015}\natexlab{}.
\newblock \showarticletitle{CogBoost: Boosting for Fast Cost-Sensitive Graph
  Classification}.
\newblock \bibinfo{journal}{\emph{{IEEE} Trans. Knowl. Data Eng.}}
  \bibinfo{volume}{27}, \bibinfo{number}{11} (\bibinfo{year}{2015}),
  \bibinfo{pages}{2933--2946}.
\newblock
\urldef\tempurl%
\url{https://doi.org/10.1109/TKDE.2015.2391115}
\showDOI{\tempurl}


\bibitem[\protect\citeauthoryear{Rieck, Bock, and Borgwardt}{Rieck
  et~al\mbox{.}}{2019}]%
        {DBLP:conf/icml/RieckBB19}
\bibfield{author}{\bibinfo{person}{Bastian Rieck}, \bibinfo{person}{Christian
  Bock}, {and} \bibinfo{person}{Karsten~M. Borgwardt}.}
  \bibinfo{year}{2019}\natexlab{}.
\newblock \showarticletitle{A Persistent Weisfeiler-Lehman Procedure for Graph
  Classification}. In \bibinfo{booktitle}{\emph{ICML}}.
\newblock


\bibitem[\protect\citeauthoryear{Rocco, Arandjelovic, and Sivic}{Rocco
  et~al\mbox{.}}{2019}]%
        {DBLP:journals/pami/RoccoAS19}
\bibfield{author}{\bibinfo{person}{Ignacio Rocco}, \bibinfo{person}{Relja
  Arandjelovic}, {and} \bibinfo{person}{Josef Sivic}.}
  \bibinfo{year}{2019}\natexlab{}.
\newblock \showarticletitle{Convolutional Neural Network Architecture for
  Geometric Matching}.
\newblock \bibinfo{journal}{\emph{{IEEE} Trans. Pattern Anal. Mach. Intell.}}
  \bibinfo{volume}{41}, \bibinfo{number}{11} (\bibinfo{year}{2019}).
\newblock


\bibitem[\protect\citeauthoryear{Rong, Huang, Xu, and Huang}{Rong
  et~al\mbox{.}}{2020}]%
        {DBLP:conf/iclr/RongHXH20}
\bibfield{author}{\bibinfo{person}{Yu Rong}, \bibinfo{person}{Wenbing Huang},
  \bibinfo{person}{Tingyang Xu}, {and} \bibinfo{person}{Junzhou Huang}.}
  \bibinfo{year}{2020}\natexlab{}.
\newblock \showarticletitle{DropEdge: Towards Deep Graph Convolutional Networks
  on Node Classification}. In \bibinfo{booktitle}{\emph{ICLR}}.
\newblock


\bibitem[\protect\citeauthoryear{Shervashidze, Schweitzer, van Leeuwen,
  Mehlhorn, and Borgwardt}{Shervashidze et~al\mbox{.}}{2011}]%
        {DBLP:journals/jmlr/ShervashidzeSLMB11}
\bibfield{author}{\bibinfo{person}{Nino Shervashidze}, \bibinfo{person}{Pascal
  Schweitzer}, \bibinfo{person}{Erik~Jan van Leeuwen}, \bibinfo{person}{Kurt
  Mehlhorn}, {and} \bibinfo{person}{Karsten~M. Borgwardt}.}
  \bibinfo{year}{2011}\natexlab{}.
\newblock \showarticletitle{Weisfeiler-Lehman Graph Kernels}.
\newblock \bibinfo{journal}{\emph{J. Mach. Learn. Res.}}  \bibinfo{volume}{12}
  (\bibinfo{year}{2011}).
\newblock


\bibitem[\protect\citeauthoryear{Shorten and Khoshgoftaar}{Shorten and
  Khoshgoftaar}{2019}]%
        {DBLP:journals/jbd/ShortenK19}
\bibfield{author}{\bibinfo{person}{Connor Shorten} {and}
  \bibinfo{person}{Taghi~M. Khoshgoftaar}.} \bibinfo{year}{2019}\natexlab{}.
\newblock \showarticletitle{A survey on Image Data Augmentation for Deep
  Learning}.
\newblock \bibinfo{journal}{\emph{J. Big Data}}  \bibinfo{volume}{6}
  (\bibinfo{year}{2019}), \bibinfo{pages}{60}.
\newblock


\bibitem[\protect\citeauthoryear{Singh and Lee}{Singh and Lee}{2017}]%
        {DBLP:conf/iccv/SinghL17}
\bibfield{author}{\bibinfo{person}{Krishna~Kumar Singh} {and}
  \bibinfo{person}{Yong~Jae Lee}.} \bibinfo{year}{2017}\natexlab{}.
\newblock \showarticletitle{Hide-and-Seek: Forcing a Network to be Meticulous
  for Weakly-Supervised Object and Action Localization}. In
  \bibinfo{booktitle}{\emph{ICCV}}.
\newblock


\bibitem[\protect\citeauthoryear{Takahashi, Matsubara, and Uehara}{Takahashi
  et~al\mbox{.}}{2020}]%
        {DBLP:journals/tcsv/TakahashiMU20}
\bibfield{author}{\bibinfo{person}{Ryo Takahashi}, \bibinfo{person}{Takashi
  Matsubara}, {and} \bibinfo{person}{Kuniaki Uehara}.}
  \bibinfo{year}{2020}\natexlab{}.
\newblock \showarticletitle{Data Augmentation Using Random Image Cropping and
  Patching for Deep CNNs}.
\newblock \bibinfo{journal}{\emph{{IEEE} Trans. Circuits Syst. Video Technol.}}
  \bibinfo{volume}{30}, \bibinfo{number}{9} (\bibinfo{year}{2020}).
\newblock


\bibitem[\protect\citeauthoryear{Van~der Maaten and Hinton}{Van~der Maaten and
  Hinton}{2008}]%
        {van2008visualizing}
\bibfield{author}{\bibinfo{person}{Laurens Van~der Maaten} {and}
  \bibinfo{person}{Geoffrey Hinton}.} \bibinfo{year}{2008}\natexlab{}.
\newblock \showarticletitle{Visualizing data using t-SNE.}
\newblock \bibinfo{journal}{\emph{Journal of machine learning research}}
  \bibinfo{volume}{9}, \bibinfo{number}{11} (\bibinfo{year}{2008}).
\newblock


\bibitem[\protect\citeauthoryear{Verma, Qu, Lamb, Bengio, Kannala, and
  Tang}{Verma et~al\mbox{.}}{2019}]%
        {DBLP:journals/corr/abs-1909-11715}
\bibfield{author}{\bibinfo{person}{Vikas Verma}, \bibinfo{person}{Meng Qu},
  \bibinfo{person}{Alex Lamb}, \bibinfo{person}{Yoshua Bengio},
  \bibinfo{person}{Juho Kannala}, {and} \bibinfo{person}{Jian Tang}.}
  \bibinfo{year}{2019}\natexlab{}.
\newblock \showarticletitle{GraphMix: Regularized Training of Graph Neural
  Networks for Semi-Supervised Learning}.
\newblock \bibinfo{journal}{\emph{CoRR}}  \bibinfo{volume}{abs/1909.11715}
  (\bibinfo{year}{2019}).
\newblock
\showeprint[arxiv]{1909.11715}


\bibitem[\protect\citeauthoryear{Wang, Wang, Liang, Cai, and Hooi}{Wang
  et~al\mbox{.}}{2020a}]%
        {DBLP:journals/corr/abs-2009-10564}
\bibfield{author}{\bibinfo{person}{Yiwei Wang}, \bibinfo{person}{Wei Wang},
  \bibinfo{person}{Yuxuan Liang}, \bibinfo{person}{Yujun Cai}, {and}
  \bibinfo{person}{Bryan Hooi}.} \bibinfo{year}{2020}\natexlab{a}.
\newblock \showarticletitle{GraphCrop: Subgraph Cropping for Graph
  Classification}.
\newblock \bibinfo{journal}{\emph{CoRR}}  \bibinfo{volume}{abs/2009.10564}
  (\bibinfo{year}{2020}).
\newblock
\showeprint[arxiv]{2009.10564}


\bibitem[\protect\citeauthoryear{Wang, Wang, Liang, Cai, and Hooi}{Wang
  et~al\mbox{.}}{2021a}]%
        {DBLP:conf/www/Wang0LCH21}
\bibfield{author}{\bibinfo{person}{Yiwei Wang}, \bibinfo{person}{Wei Wang},
  \bibinfo{person}{Yuxuan Liang}, \bibinfo{person}{Yujun Cai}, {and}
  \bibinfo{person}{Bryan Hooi}.} \bibinfo{year}{2021}\natexlab{a}.
\newblock \showarticletitle{CurGraph: Curriculum Learning for Graph
  Classification}. In \bibinfo{booktitle}{\emph{WWW}}.
\newblock


\bibitem[\protect\citeauthoryear{Wang, Wang, Liang, Cai, and Hooi}{Wang
  et~al\mbox{.}}{2021b}]%
        {DBLP:conf/www/WangWLCH21}
\bibfield{author}{\bibinfo{person}{Yiwei Wang}, \bibinfo{person}{Wei Wang},
  \bibinfo{person}{Yuxuan Liang}, \bibinfo{person}{Yujun Cai}, {and}
  \bibinfo{person}{Bryan Hooi}.} \bibinfo{year}{2021}\natexlab{b}.
\newblock \showarticletitle{Mixup for Node and Graph Classification}. In
  \bibinfo{booktitle}{\emph{WWW}}.
\newblock


\bibitem[\protect\citeauthoryear{Wang, Wang, Liang, Cai, Liu, and Hooi}{Wang
  et~al\mbox{.}}{2020b}]%
        {DBLP:conf/kdd/WangWLCLH20}
\bibfield{author}{\bibinfo{person}{Yiwei Wang}, \bibinfo{person}{Wei Wang},
  \bibinfo{person}{Yuxuan Liang}, \bibinfo{person}{Yujun Cai},
  \bibinfo{person}{Juncheng Liu}, {and} \bibinfo{person}{Bryan Hooi}.}
  \bibinfo{year}{2020}\natexlab{b}.
\newblock \showarticletitle{NodeAug: Semi-Supervised Node Classification with
  Data Augmentation}. In \bibinfo{booktitle}{\emph{KDD}}.
\newblock


\bibitem[\protect\citeauthoryear{Wen, Sun, Yang, Song, Gao, Wang, and Xu}{Wen
  et~al\mbox{.}}{2021}]%
        {DBLP:conf/ijcai/Wen0YSGWX21}
\bibfield{author}{\bibinfo{person}{Qingsong Wen}, \bibinfo{person}{Liang Sun},
  \bibinfo{person}{Fan Yang}, \bibinfo{person}{Xiaomin Song},
  \bibinfo{person}{Jingkun Gao}, \bibinfo{person}{Xue Wang}, {and}
  \bibinfo{person}{Huan Xu}.} \bibinfo{year}{2021}\natexlab{}.
\newblock \showarticletitle{Time Series Data Augmentation for Deep Learning:
  {A} Survey}. In \bibinfo{booktitle}{\emph{IJCAI}}.
\newblock


\bibitem[\protect\citeauthoryear{Xu, Chen, Liu, Chen, Weng, Hong, and Lin}{Xu
  et~al\mbox{.}}{2019a}]%
        {DBLP:conf/ijcai/XuC0CWHL19}
\bibfield{author}{\bibinfo{person}{Kaidi Xu}, \bibinfo{person}{Hongge Chen},
  \bibinfo{person}{Sijia Liu}, \bibinfo{person}{Pin{-}Yu Chen},
  \bibinfo{person}{Tsui{-}Wei Weng}, \bibinfo{person}{Mingyi Hong}, {and}
  \bibinfo{person}{Xue Lin}.} \bibinfo{year}{2019}\natexlab{a}.
\newblock \showarticletitle{Topology Attack and Defense for Graph Neural
  Networks: An Optimization Perspective}. In \bibinfo{booktitle}{\emph{IJCAI}}.
\newblock


\bibitem[\protect\citeauthoryear{Xu, Hu, Leskovec, and Jegelka}{Xu
  et~al\mbox{.}}{2019b}]%
        {DBLP:conf/iclr/XuHLJ19}
\bibfield{author}{\bibinfo{person}{Keyulu Xu}, \bibinfo{person}{Weihua Hu},
  \bibinfo{person}{Jure Leskovec}, {and} \bibinfo{person}{Stefanie Jegelka}.}
  \bibinfo{year}{2019}\natexlab{b}.
\newblock \showarticletitle{How Powerful are Graph Neural Networks?}. In
  \bibinfo{booktitle}{\emph{ICLR}}.
\newblock


\bibitem[\protect\citeauthoryear{Yanardag and Vishwanathan}{Yanardag and
  Vishwanathan}{2015}]%
        {DBLP:conf/kdd/YanardagV15}
\bibfield{author}{\bibinfo{person}{Pinar Yanardag} {and}
  \bibinfo{person}{S.~V.~N. Vishwanathan}.} \bibinfo{year}{2015}\natexlab{}.
\newblock \showarticletitle{Deep Graph Kernels}. In
  \bibinfo{booktitle}{\emph{KDD}}.
\newblock


\bibitem[\protect\citeauthoryear{Ying, He, Chen, Eksombatchai, Hamilton, and
  Leskovec}{Ying et~al\mbox{.}}{2018a}]%
        {DBLP:conf/kdd/YingHCEHL18}
\bibfield{author}{\bibinfo{person}{Rex Ying}, \bibinfo{person}{Ruining He},
  \bibinfo{person}{Kaifeng Chen}, \bibinfo{person}{Pong Eksombatchai},
  \bibinfo{person}{William~L. Hamilton}, {and} \bibinfo{person}{Jure
  Leskovec}.} \bibinfo{year}{2018}\natexlab{a}.
\newblock \showarticletitle{Graph Convolutional Neural Networks for Web-Scale
  Recommender Systems}. In \bibinfo{booktitle}{\emph{KDD}}.
\newblock


\bibitem[\protect\citeauthoryear{Ying, You, Morris, Ren, Hamilton, and
  Leskovec}{Ying et~al\mbox{.}}{2018b}]%
        {DBLP:conf/nips/YingY0RHL18}
\bibfield{author}{\bibinfo{person}{Zhitao Ying}, \bibinfo{person}{Jiaxuan You},
  \bibinfo{person}{Christopher Morris}, \bibinfo{person}{Xiang Ren},
  \bibinfo{person}{William~L. Hamilton}, {and} \bibinfo{person}{Jure
  Leskovec}.} \bibinfo{year}{2018}\natexlab{b}.
\newblock \showarticletitle{Hierarchical Graph Representation Learning with
  Differentiable Pooling}. In \bibinfo{booktitle}{\emph{NeurIPS}}.
\newblock


\bibitem[\protect\citeauthoryear{Yoo, Jeon, and Kang}{Yoo
  et~al\mbox{.}}{2019}]%
        {DBLP:conf/ijcai/YooJK19}
\bibfield{author}{\bibinfo{person}{Jaemin Yoo}, \bibinfo{person}{Hyunsik Jeon},
  {and} \bibinfo{person}{U Kang}.} \bibinfo{year}{2019}\natexlab{}.
\newblock \showarticletitle{Belief Propagation Network for Hard Inductive
  Semi-Supervised Learning}. In \bibinfo{booktitle}{\emph{IJCAI}}.
\newblock


\bibitem[\protect\citeauthoryear{Yoo, Kang, Scanagatta, Corani, and
  Zaffalon}{Yoo et~al\mbox{.}}{2020}]%
        {DBLP:conf/wsdm/YooKSCZ20}
\bibfield{author}{\bibinfo{person}{Jaemin Yoo}, \bibinfo{person}{U Kang},
  \bibinfo{person}{Mauro Scanagatta}, \bibinfo{person}{Giorgio Corani}, {and}
  \bibinfo{person}{Marco Zaffalon}.} \bibinfo{year}{2020}\natexlab{}.
\newblock \showarticletitle{Sampling Subgraphs with Guaranteed Treewidth for
  Accurate and Efficient Graphical Inference}. In
  \bibinfo{booktitle}{\emph{WSDM}}.
\newblock


\bibitem[\protect\citeauthoryear{Yoo, Kim, Yoon, Kim, Jang, and Kang}{Yoo
  et~al\mbox{.}}{2021}]%
        {DBLP:conf/icdm/YooKYKJK21}
\bibfield{author}{\bibinfo{person}{Jaemin Yoo}, \bibinfo{person}{Junghun Kim},
  \bibinfo{person}{Hoyoung Yoon}, \bibinfo{person}{Geonsoo Kim},
  \bibinfo{person}{Changwon Jang}, {and} \bibinfo{person}{U Kang}.}
  \bibinfo{year}{2021}\natexlab{}.
\newblock \showarticletitle{Accurate Graph-Based {PU} Learning without Class
  Prior}. In \bibinfo{booktitle}{\emph{ICDM}}.
\newblock


\bibitem[\protect\citeauthoryear{You, Chen, Shen, and Wang}{You
  et~al\mbox{.}}{2021}]%
        {DBLP:conf/icml/YouCSW21}
\bibfield{author}{\bibinfo{person}{Yuning You}, \bibinfo{person}{Tianlong
  Chen}, \bibinfo{person}{Yang Shen}, {and} \bibinfo{person}{Zhangyang Wang}.}
  \bibinfo{year}{2021}\natexlab{}.
\newblock \showarticletitle{Graph Contrastive Learning Automated}. In
  \bibinfo{booktitle}{\emph{ICML}}.
\newblock


\bibitem[\protect\citeauthoryear{You, Chen, Sui, Chen, Wang, and Shen}{You
  et~al\mbox{.}}{2020}]%
        {DBLP:conf/nips/YouCSCWS20}
\bibfield{author}{\bibinfo{person}{Yuning You}, \bibinfo{person}{Tianlong
  Chen}, \bibinfo{person}{Yongduo Sui}, \bibinfo{person}{Ting Chen},
  \bibinfo{person}{Zhangyang Wang}, {and} \bibinfo{person}{Yang Shen}.}
  \bibinfo{year}{2020}\natexlab{}.
\newblock \showarticletitle{Graph Contrastive Learning with Augmentations}. In
  \bibinfo{booktitle}{\emph{NeurIPS}}.
\newblock


\bibitem[\protect\citeauthoryear{Yun, Han, Chun, Oh, Yoo, and Choe}{Yun
  et~al\mbox{.}}{2019}]%
        {DBLP:conf/iccv/YunHCOYC19}
\bibfield{author}{\bibinfo{person}{Sangdoo Yun}, \bibinfo{person}{Dongyoon
  Han}, \bibinfo{person}{Sanghyuk Chun}, \bibinfo{person}{Seong~Joon Oh},
  \bibinfo{person}{Youngjoon Yoo}, {and} \bibinfo{person}{Junsuk Choe}.}
  \bibinfo{year}{2019}\natexlab{}.
\newblock \showarticletitle{CutMix: Regularization Strategy to Train Strong
  Classifiers With Localizable Features}. In \bibinfo{booktitle}{\emph{ICCV}}.
\newblock


\bibitem[\protect\citeauthoryear{Zeng, Zhou, Srivastava, Kannan, and
  Prasanna}{Zeng et~al\mbox{.}}{2020}]%
        {DBLP:conf/iclr/ZengZSKP20}
\bibfield{author}{\bibinfo{person}{Hanqing Zeng}, \bibinfo{person}{Hongkuan
  Zhou}, \bibinfo{person}{Ajitesh Srivastava}, \bibinfo{person}{Rajgopal
  Kannan}, {and} \bibinfo{person}{Viktor~K. Prasanna}.}
  \bibinfo{year}{2020}\natexlab{}.
\newblock \showarticletitle{GraphSAINT: Graph Sampling Based Inductive Learning
  Method}. In \bibinfo{booktitle}{\emph{ICLR}}.
\newblock


\bibitem[\protect\citeauthoryear{Zhang, Ciss{\'{e}}, Dauphin, and
  Lopez{-}Paz}{Zhang et~al\mbox{.}}{2018}]%
        {DBLP:conf/iclr/ZhangCDL18}
\bibfield{author}{\bibinfo{person}{Hongyi Zhang}, \bibinfo{person}{Moustapha
  Ciss{\'{e}}}, \bibinfo{person}{Yann~N. Dauphin}, {and} \bibinfo{person}{David
  Lopez{-}Paz}.} \bibinfo{year}{2018}\natexlab{}.
\newblock \showarticletitle{mixup: Beyond Empirical Risk Minimization}. In
  \bibinfo{booktitle}{\emph{ICLR}}.
\newblock


\bibitem[\protect\citeauthoryear{Zhao, Liu, Neves, Woodford, Jiang, and
  Shah}{Zhao et~al\mbox{.}}{2020}]%
        {DBLP:journals/corr/abs-2006-06830}
\bibfield{author}{\bibinfo{person}{Tong Zhao}, \bibinfo{person}{Yozen Liu},
  \bibinfo{person}{Leonardo Neves}, \bibinfo{person}{Oliver~J. Woodford},
  \bibinfo{person}{Meng Jiang}, {and} \bibinfo{person}{Neil Shah}.}
  \bibinfo{year}{2020}\natexlab{}.
\newblock \showarticletitle{Data Augmentation for Graph Neural Networks}.
\newblock \bibinfo{journal}{\emph{CoRR}}  \bibinfo{volume}{abs/2006.06830}
  (\bibinfo{year}{2020}).
\newblock
\showeprint[arxiv]{2006.06830}


\bibitem[\protect\citeauthoryear{Zhong, Zheng, Kang, Li, and Yang}{Zhong
  et~al\mbox{.}}{2020}]%
        {DBLP:conf/aaai/Zhong0KL020}
\bibfield{author}{\bibinfo{person}{Zhun Zhong}, \bibinfo{person}{Liang Zheng},
  \bibinfo{person}{Guoliang Kang}, \bibinfo{person}{Shaozi Li}, {and}
  \bibinfo{person}{Yi Yang}.} \bibinfo{year}{2020}\natexlab{}.
\newblock \showarticletitle{Random Erasing Data Augmentation}. In
  \bibinfo{booktitle}{\emph{AAAI}}.
\newblock


\bibitem[\protect\citeauthoryear{Zhou, Shen, and Xuan}{Zhou
  et~al\mbox{.}}{2020}]%
        {DBLP:conf/cikm/ZhouSX20}
\bibfield{author}{\bibinfo{person}{Jiajun Zhou}, \bibinfo{person}{Jie Shen},
  {and} \bibinfo{person}{Qi Xuan}.} \bibinfo{year}{2020}\natexlab{}.
\newblock \showarticletitle{Data Augmentation for Graph Classification}. In
  \bibinfo{booktitle}{\emph{CIKM}}.
\newblock


\bibitem[\protect\citeauthoryear{Zhu, Liu, Li, Wan, and Qin}{Zhu
  et~al\mbox{.}}{2018}]%
        {DBLP:conf/pakdd/ZhuLLWQ18}
\bibfield{author}{\bibinfo{person}{Xinyue Zhu}, \bibinfo{person}{Yifan Liu},
  \bibinfo{person}{Jiahong Li}, \bibinfo{person}{Tao Wan}, {and}
  \bibinfo{person}{Zengchang Qin}.} \bibinfo{year}{2018}\natexlab{}.
\newblock \showarticletitle{Emotion Classification with Data Augmentation Using
  Generative Adversarial Networks}. In \bibinfo{booktitle}{\emph{PAKDD}}.
\newblock


\bibitem[\protect\citeauthoryear{Z{\"{u}}gner and G{\"{u}}nnemann}{Z{\"{u}}gner
  and G{\"{u}}nnemann}{2019}]%
        {DBLP:conf/iclr/ZugnerG19}
\bibfield{author}{\bibinfo{person}{Daniel Z{\"{u}}gner} {and}
  \bibinfo{person}{Stephan G{\"{u}}nnemann}.} \bibinfo{year}{2019}\natexlab{}.
\newblock \showarticletitle{Adversarial Attacks on Graph Neural Networks via
  Meta Learning}. In \bibinfo{booktitle}{\emph{ICLR}}.
\newblock


\end{thebibliography}

\clearpage
\appendix
\section{Unbiasedness of \methodone}
\label{appendix:proof}

\subsection{Properties of Intermediate Graphs}

We have four graphs $G$, $G'$, $G''$, and $\bar{G}$ presented in Algorithm \ref{alg:method-one}, which denote the original graph, the graph after the split, the graph after the adjustment, and the final graph generated by \methodone, respectively.
Let $v_i$ be the target node of the split operation, $d_i$ be the degree of $v_i$ in $G$, and $h_i$ be the \emph{expected} number of edges added in the adjustment operation.
Let $T(\cdot)$ be the function that counts the number of triangles in a graph.
Then, we analyze the properties of graphs generated during \methodone in Lemma \ref{lemma:triangles-1} to \ref{lemma:triangles-4}.

%\begin{lemma}
%	$\mathbb{E}_{u \in \mathcal{V}}[t_u] = 3T(G) / |\mathcal{V}|$ for any node $u \in \mathcal{V}$.
%\label{lemma:triangles-0}
%\end{lemma}
%
%\begin{proof}
%	The proof is straightforward due to the following relation that holds for any graph and node: $\sum_{v \in \mathcal{V}} t_v = 3T(G)$.
%\end{proof}

\begin{lemma}
	$\mathbb{E}[T(G')] = T(G) - t_i / 2$.
\label{lemma:triangles-1}
\end{lemma}

\begin{proof}
\vspace{-1mm}
	Each triangle in $G$ containing $v_i$ has a probability of $0.5$ to be removed in the split operation, due to the random selection of the target node from $v_j$ and $v_k$.
	Thus, it is expected that the half of all triangles containing $v_i$ are removed in $G'$.
\vspace{-1mm}
\end{proof}

\begin{lemma}
	In the adjustment operation of Algorithm \ref{alg:method-one-adjust}, let $l$ be the number of new triangles created in $G''$ by connecting an edge between a target node $u \in \mathcal{S}$ and either $v_j$ or $v_k$.
	Then, $\mathbb{E}[l] = |\mathcal{N}_{ui}| / 2 + 1$, where $\mathcal{N}_{ui}$ is the set of common neighbors between $u$ and $v_i$ in $G$.
\label{lemma:triangles-2}
\end{lemma}

\begin{proof}
\vspace{-1mm}
	Node $u$ is connected to either $v_j$ or $v_k$ in $G'$, since it is a neighbor of $v_i$ in the the original graph $G$ before the split is done.
	Assume that $u$ is connected to $v_j$ in $G'$ without loss of generality.
	Then, the adjustment for $u$ makes an edge $(u, v_k)$.
	A single triangle $(u, v_j, v_k)$ is always created by this operation, since edges $(u, v_j)$ and $(v_j, v_k)$ already exist in $G'$.
	The number of additional triangles created by the adjustment is the same as the number of common neighbors between $u$ and $v_k$ in $G'$, denoted by $|\mathcal{N}_{uk}'|$.
	Since we split $\mathcal{N}_i$ by the same probability into $v_j$ and $v_k$ during the split operation, $\mathbb{E}[|\mathcal{N}_{uk}'|] = |\mathcal{N}_{ui}|/2$.
	We prove the lemma by adding one.
\vspace{-1mm}
\end{proof}

\begin{lemma}
	$\mathbb{E}[T(G'')] = T(G) + h_i(t_i / d_i + 1) - t_i / 2$.
\label{lemma:triangles-3}
\end{lemma}

\begin{proof}
\vspace{-1mm}
	Following Lemma \ref{lemma:triangles-2}, $|\mathcal{N}_{ui}| / 2 + 1$ triangles are expected to be created for each target node $u \in \mathcal{S}$ during the adjustment.
	We estimate $|\mathcal{N}_{ui}|$ by $2t_i/d_i$ based on the relation $$\sum_{k \in \mathcal{N}_i} |\mathcal{N}_{ki}| = 2t_i,$$ where $\mathcal{N}_i$ is the set of neighbors of $v_i$ in $G$.
	We get $$\mathbb{E}[T(G'') - T(G')] = h_i(t_i/d_i + 1)$$ by repeating the adjustment $h_i$ times for every possible $u \in \mathcal{S}$.
	We prove the lemma by combining it with Lemma \ref{lemma:triangles-1}.
\vspace{-1mm}
\end{proof}

\begin{lemma}
	$\mathbb{E}[|\bar{\mathcal{E}}|] = |\mathcal{E}''| - 3T(G'') / |\mathcal{E}''| - 1$.
\label{lemma:triangles-4}
\end{lemma}

\begin{proof}
\vspace{-1mm}
	Let $v_o$ and $v_p$ be the target nodes of the merge operation, and let $t_{op}''$ be the number of triangles containing $v_o$ and $v_p$ in $G''$.
	A single edge $(v_o, v_p)$ is always removed by the merge operation as $v_o$ and $v_p$ are merged into one.
	The number of additional edges removed by the merge operation is given as $t_{op}''$, which is estimated by $3T(G'') / |\mathcal{E}''|$ based on the relation $$\sum_{(i, j) \in \mathcal{E}''} t''_{ij} = 3T(G'').$$
	We prove the lemma by adding one to $3T(G'') / |\mathcal{E}''|$.
\vspace{-1mm}
\end{proof}

\subsection{Proof of Lemma \ref{lemma:adjustment}}

We restate Lemma \ref{lemma:adjustment} given in Section \ref{sssec:nodesam-properties} and present the full proof based on Lemma \ref{lemma:triangles-1} to \ref{lemma:triangles-4}.

% TODO: To align the number with the original lemma.
\begin{customlma}{1}
	Given a graph $G = (\mathcal{V}, \mathcal{E}, \mathbf{X})$, let $\bar{G} = (\bar{\mathcal{V}}, \bar{\mathcal{E}}, \bar{\mathbf{X}})$ be the result of $\methodone$.
	Then, $\mathbb{E}[|\bar{\mathcal{V}}| - |\mathcal{V}|] = 0$ and $\mathbb{E}[|\bar{\mathcal{E}}| - |\mathcal{E}|] = 0$.
\label{lemma:adjustment}
\end{customlma}

\begin{proof}
\vspace{-1mm}
	We prove only the edge part, since it is straightforward that \methodone does not change the number of nodes.
	Let $m$ be the expected number of edges removed by the merge operation:
	\begin{equation}
		m \equiv \mathbb{E}[|\mathcal{E}''| - |\bar{\mathcal{E}}|].
	\label{eq:triangle-0.2}
	\end{equation}

	Then, $m$ is rewritten as follows by Lemma \ref{lemma:triangles-4}:
	\begin{equation}
		m = \mathbb{E}[3T(G'') / |\mathcal{E}''|] + 1.
	\label{eq:triangle-0.5}
	\end{equation}
	
	If we ignore the dependence between $T(G'')$ and $|\mathcal{E}''|$, which is negligible in real-world graphs that sastify $|\mathcal{E}| \gg h_i$, Equation \eqref{eq:triangle-0.5} changes into
	\begin{equation}
		m = \frac{3\mathbb{E}[T(G'')]}{\mathbb{E}[|\mathcal{E}''|]} + 1.
	\label{eq:triangle-1}
	\end{equation}

	The first term $\mathbb{E}[T(G'')]$ is rewritten by Lemma \ref{lemma:triangles-3}:
	\begin{equation}
		\mathbb{E}[T(G'')]
			= |\mathcal{V}| t_i/3 + h_i(t_i / d_i + 1) - t_i / 2,
	\label{eq:triangle-2}
	\end{equation}
	where the global number $T(G)$ of triangles is replaced with a local estimation $|\mathcal{V}| t_i/3$, since we select $v_i$ from $G$ uniformly at random and the relation $\mathbb{E}_i [t_i] = 3T(G)$ holds between $T(G)$ and $t_i$.

%	\begin{equation}
%	\begin{split}
%		\mathbb{E}[T(G'')]
%			&= T(G) + h_i(t_i / d_i + 1) - t_i / 2 \\
%			&\approx |\mathcal{V}| t_i/3 + h_i(t_i / d_i + 1) - t_i / 2.
%	\end{split}
%	\label{eq:triangle-2}
%	\end{equation}
%	We use the relation $\sum_{v \in \mathcal{V}} t_v = 3T(G)$ to make the approximation of the global number of triangles.
%	Still, the approximation does not harm the generality of our proof, since we sample the target $v_i$ of the split operation uniformly at random, i.e., $\mathbb{E}_i [t_i] = 3T(G)$.

	The second term $\mathbb{E}[|\mathcal{E}''|]$ is given by the definition of $h_i$:
	\begin{equation}
		\mathbb{E}[|\mathcal{E}''|] = |\mathcal{E}| + h_i + 1.
	\label{eq:triangle-3}
	\end{equation}
	
	We apply Equation \eqref{eq:triangle-2} and \eqref{eq:triangle-3} to Equation \eqref{eq:triangle-1} to get
	\begin{equation}
		m = \frac{3(t_i/d_i + 1)h_i + (|\mathcal{V}| - 3/2)t_i}{h_i + |\mathcal{E}| + 1} + 1.
	\label{eq:triangle-4}
	\end{equation}

%	Equation \eqref{eq:adjustment} is the solution of the following quadratic equation with respect to $h_i$:
%	\begin{equation}
%		h_i = \frac{3(t_i/d_i + 1)h_i + (|\mathcal{V}| - 3/2)t_i}{h_i + |\mathcal{E}| + 1}.
%	\end{equation}
	
	If we denote the first term of the right hand side of Equation \eqref{eq:triangle-4} as $A$, we can verify that Equation \eqref{eq:adjustment} is the solution of $h_i = A$:
	\begin{equation}
	\begin{split}
		&h_i = A \\
		&h_i (h_i + |\mathcal{E}| + 1) = 3(t_i/d_i + 1)h_i + (|\mathcal{V}| - 3/2)t_i \\
		&h_i^2 + (|\mathcal{E}| - 3t_i/d_i - 2)h_i - (|\mathcal{V}| - 3/2)t_i = 0.
		\end{split}
	\end{equation}
	
	Then, the right hand side of Equation \eqref{eq:triangle-4} changes into $h_i + 1$.
	By the definition of $m$ and $h_i$, we finally get
	\begin{equation*}
		\mathbb{E}[|\mathcal{E}''| - |\bar{\mathcal{E}}|]
			= \mathbb{E}[|\mathcal{E}''|] - |\mathcal{E}| - 1 + 1,
	\end{equation*}
	which changes into $\mathbb{E}[|\bar{\mathcal{E}}| - |\mathcal{E}|] = 0$ and proves the lemma.
\vspace{-1mm}
\end{proof}

\section{Proofs of Lemma \ref{lemma:connectivity} to \ref{lemma:method-two-scalability}}

\subsection{Proof of Lemma \ref{lemma:connectivity}}
\label{appendix:proof-lemma-2}

\begin{proof}
	The split and merge operations preserve the connectivity, since they are transformations between a single node and a pair of adjacent nodes.
	The adjustment also preserves the connectivity since it makes new edges only between two-hop neighbors.
\end{proof}

\subsection{Proof of Lemma \ref{lemma:complexity-1}}
\label{appendix:proof-lemma-3}

\begin{proof}
	The time complexity of the split operation is $O(d|\mathcal{V}| + |\mathcal{E}|)$, and the complexity of the merge operation is the same.
	The time complexity of the adjustment operation is $O(\mathcal{V}| + |\mathcal{E}|)$ since it does not change $\mathbf{X}$.
	The space complexities are always smaller than or equal to the time complexities \cite{DBLP:journals/jcss/Book74}.
	We prove the lemma by combining the complexities of all these three operations.
\end{proof}

\subsection{Proof of Lemma \ref{lemma:submix-unbiasedness}}
\label{appendix:proof-lemma-5}

\begin{proof}
	The proof is straightforward for $\mathcal{V}$, since \methodtwo does not change the number of nodes.
	For $\mathcal{E}$, let $\mathcal{E}_s$ and $\mathcal{E}'_s$ be the edges of induced subgraphs of $S$ and $S'$ in graphs $G$ and $G'$, respectively.
	Then, the following equations hold, since a) \methodtwo replaces only the edges in $\mathcal{E}_s$, b) $G$ and $G'$ are selected independently, and c) $S$ and $S'$ are selected by the same diffusion algorithm:
	\begin{align*}
		\mathbb{E}_{G, \bar{G}}[|\mathcal{E}| - |\bar{\mathcal{E}}|]
			&= \mathbb{E}_{G, G'}[|\mathcal{E}_s| - |\mathcal{E}_s'|] \\
			&= \mathbb{E}_G[|\mathcal{E}_s|] - \mathbb{E}_{G'}[|\mathcal{E}_s'|] = 0.
	\end{align*}
	The fact that $\mathbb{E}[|\mathcal{E}| - |\bar{\mathcal{E}}|] = 0$ proves the lemma.
\end{proof}

\subsection{Proof of Lemma \ref{lemma:method-two-connectivity}}
\label{appendix:proof-lemma-6}

\begin{proof}
	Let $G'$ be a graph chosen for the augmentation of $G$ by \methodtwo.
	The sets $S$ and $S'$ of nodes selected for the replacement are connected due to Lemma \ref{lemma:method-two-connectivity}.
	If we treat the induced subgraphs of $S$ and $S'$ as supernodes, the replacement of $S$ with $S'$ does not change the connectivity of $G$, proving the lemma.
\end{proof}

\subsection{Proof of Lemma \ref{lemma:method-two-scalability}}
\label{appendix:proof-lemma-7}

\begin{proof}
	The time complexity of Algorithm \ref{alg:method-two} without including the sampling function is $O(pd|\mathcal{V}| + |\mathcal{E}| + |\mathcal{E}'|)$.
	We assume that the number of labels is negligible.
	The time complexity of Algorithm \ref{alg:method-two-sample} is $O(|\mathcal{E}| + |\mathcal{E}'|)$, since the PPR diffusion is $O(|\mathcal{E}|)$ and $O(|\mathcal{E}'|)$ for $G$ and $G'$, respectively.
	The space complexities are always smaller than or equal to the time complexities \cite{DBLP:journals/jcss/Book74}.
\end{proof}

\section{Node Feature Information}
\label{appendix:data}

We describe detailed information of node features in our datasets, which are summarized in Table \ref{table:datasets}.
\begin{itemize}
	\item \textbf{Molecular graphs.}
		Each node represents an element in a chemical compound, and contains a one-hot feature representing its atomic type, such as carbon or oxygen.
	\item \textbf{Twitter.}
		Each graph represents a tweet, and each node is a keyword or a symbol appearing in a tweet.
		Every node has a one-hot feature representing its keyword or symbol.
	\item \textbf{COLLAB.}
		Since the dataset does not contain node features, we compute the degree of every node and assign a one-hot feature vector to each node based on its degree.
		Thus, the length of feature vectors is the same as the number of unique degrees in the dataset.
		The same approach was also done in previous work \cite{DBLP:conf/iclr/XuHLJ19} for graph classification.
\end{itemize}

%We preprocess node features of our datasets as follows:
%\begin{itemize}
%	\item \textbf{Molecular graphs.} Each node represents an element in a chemical compound, and has a one-hot feature vector representing its atomic type.
%		We use the original datasets without additional preprocessing.
%	\item \textbf{DBLP.} Each graph contains two types of nodes: papers and keywords.
%		We design node features as $(d+1)$-dimensional one-hot vectors, where $d$ is the number of unique keywords in the dataset.
%		The remaining element represents whether a node is a paper or not.
%	\item \textbf{COLLAB.} We compute the degree of every node and assign a one-hot feature to each node based on its degree.
%		Thus, the length of feature vectors is the same as the number of unique degrees in the dataset.
%		This was also done in previous work \cite{DBLP:conf/iclr/XuHLJ19} for graph classification.
%\end{itemize}

\begin{figure}
	\begin{subfigure}{0.236\textwidth}
		\includegraphics[trim=0 2mm 0 0, clip, width=\textwidth]{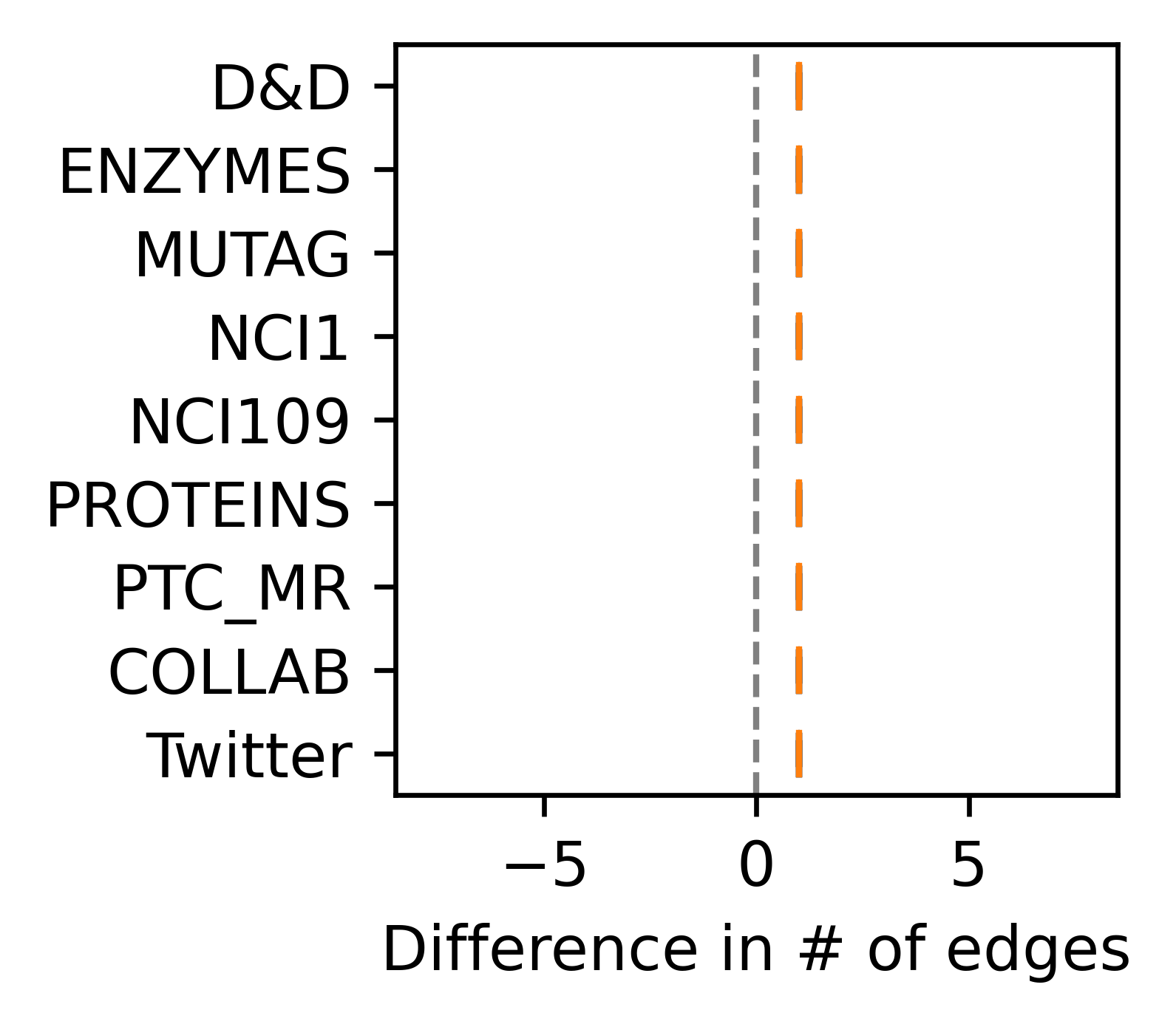}
		\caption{SplitOnly}
	\end{subfigure} \hfill
	\begin{subfigure}{0.236\textwidth}
		\includegraphics[trim=0 2mm 0 0, clip, width=\textwidth]{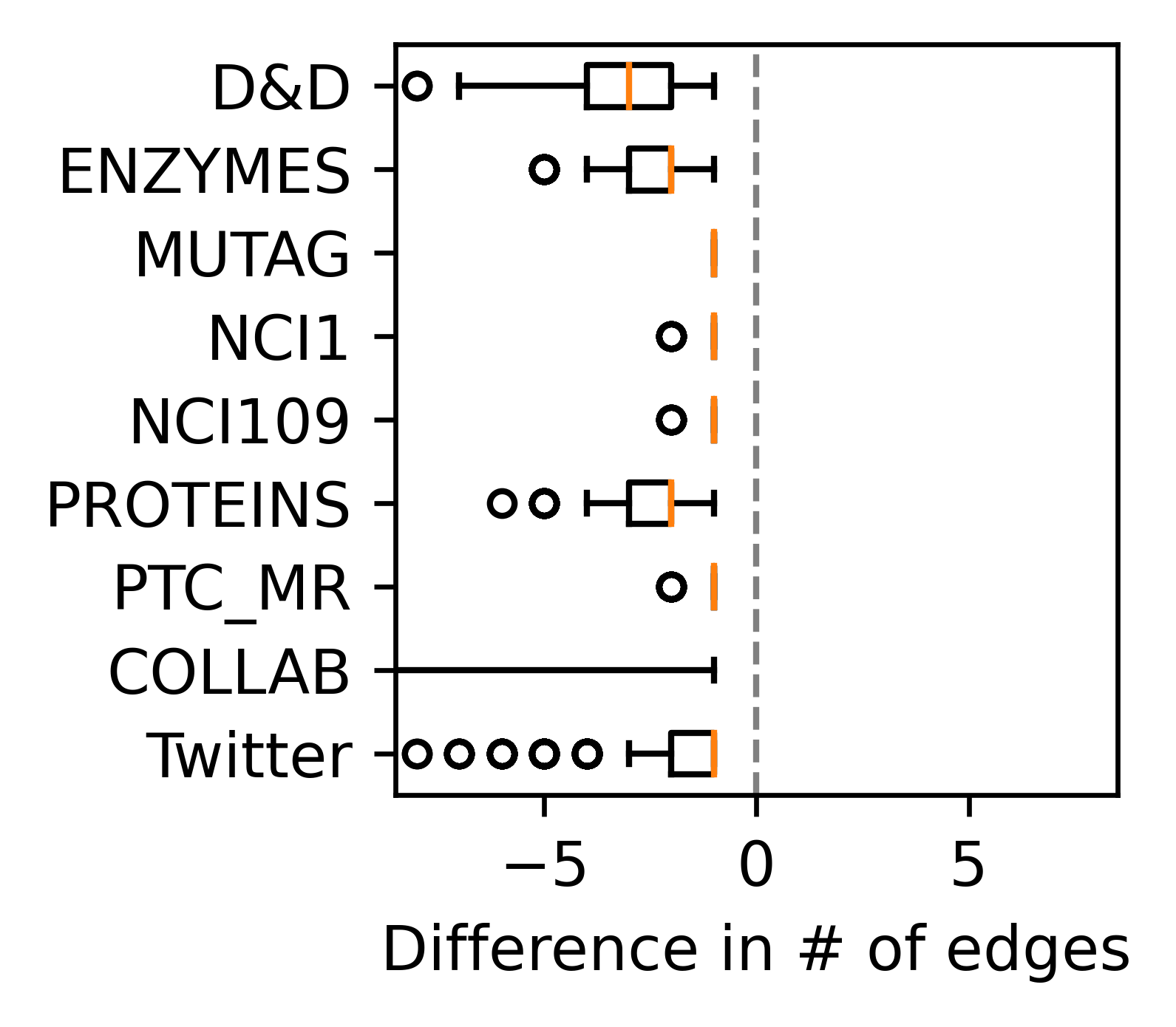}
		\caption{MergeOnly}
	\end{subfigure}
	\vspace{1mm}
	
	\begin{subfigure}{0.236\textwidth}
		\includegraphics[trim=0 2mm 0 0, clip, width=\textwidth]{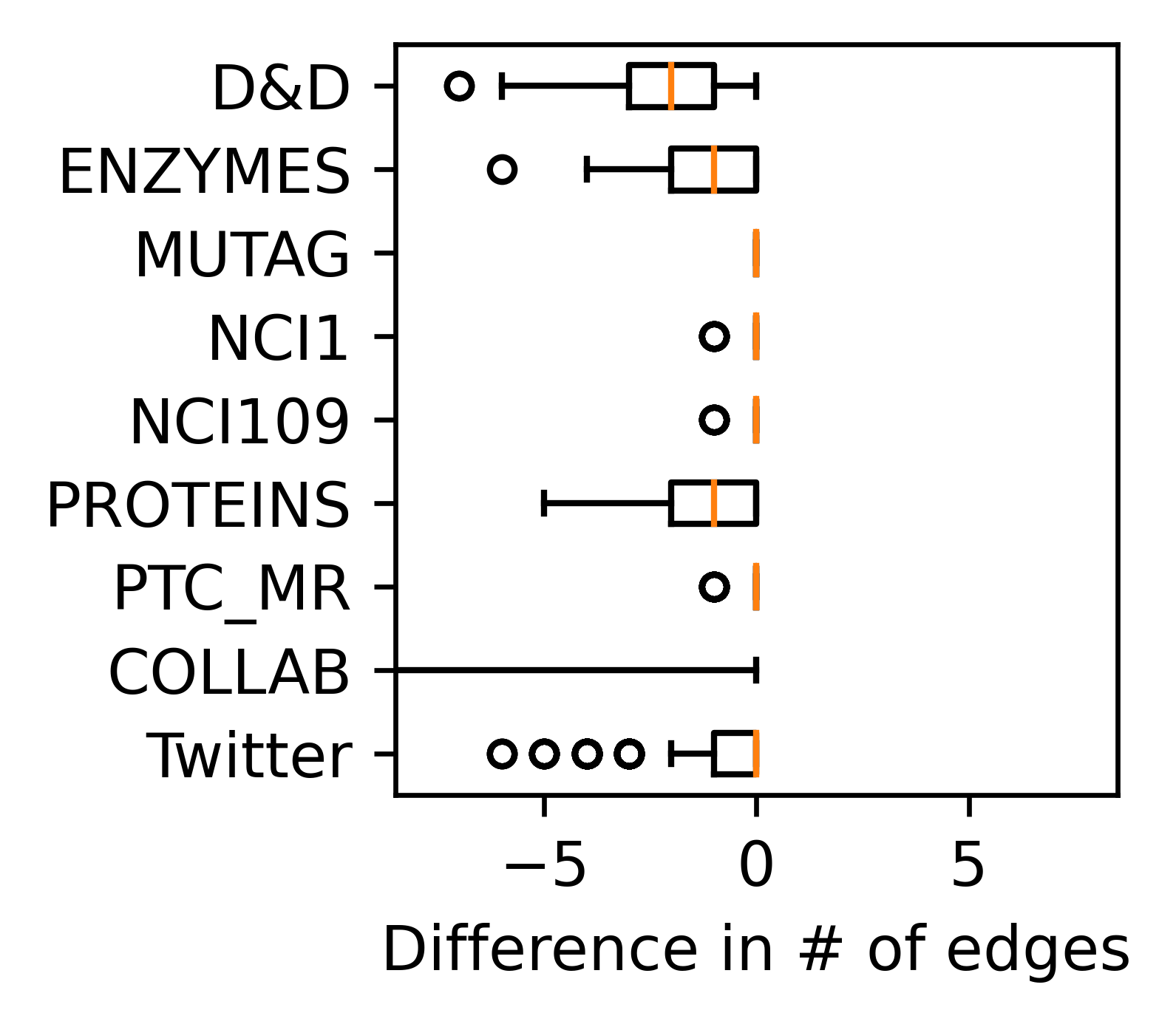}
		\caption{NodeSamBase}
	\end{subfigure} \hfill
	\begin{subfigure}{0.236\textwidth}
		\includegraphics[trim=0 2mm 0 0, clip, width=\textwidth]{figures/edges/NodeSam}
		\caption{NodeSam}
	\end{subfigure}

	\caption{
		Box plots illustrating the difference in the number of edges.
		NodeSamBase, which does not perform the adjustment operation, tends to decrease the number of edges due to the merge operation that removes additional edges.
	}
	\label{fig:edge-variance-2}
\end{figure}

\section{Ablation Study for \methodone}
\label{appendix:ablation}

We perform an additional ablation study for \methodone to confirm the effect of the adjustment operation for making unbiased changes of the number of edges.
Figure \ref{fig:edge-variance-2} shows box plots for the difference in the number of edges for all variants of \methodone: (a) SplitOnly, (b) MergeOnly, (c) NodeSamBase, and (d) \methodone.
The detailed information of such variants is discussed in Section \ref{sec:exp-ablation}.
The figure shows that the split operation increases the number of edges by one in all cases, since it splits a node into two by inserting a single edge, while the merge operation decreases the number of edges by more than one.
This is because the merge operation eliminates the triangles that contain both of the target nodes.
Our \methodone with the adjustment operation makes unbiased changes, compensating for the removed edges, compared with NodeSamBase.

\clearpage

\end{document}